\documentclass[sigconf]{acmart}
\usepackage{subfigure}
\usepackage{ragged2e}
\usepackage{algorithm}
\usepackage[noend]{algorithmic}
\newcommand{\argmin}{\operatornamewithlimits{arg\,min}}
\newcommand{\argmax}{\operatornamewithlimits{arg\,max}}
\newcommand{\KS}{{\sf KS}}
\newcommand{\CE}{{\sf CE}}
\newcommand{\CEs}{{\sf CEs}}
\newcommand{\LCE}{{\sf LCE}}
\newcommand{\GCE}{{\sf GCE}}
\newcommand{\LCEs}{{\sf LCEs}}
\newcommand{\GCEs}{{\sf GCEs}}

\newcommand{\WL}{{\sf WL}}
\newcommand{\GNN}{{\sf GNN}}
\newcommand{\GNNs}{{\sf GNNs}}
\newcommand{\LocalCEB}{{\sf LocalCE-B}}
\newcommand{\LocalCEI}{{\sf LocalCE-I}}
\newcommand{\LocalCEIwoWC}{{\sf LocalCE-I w/o WC}}
\newcommand{\LocalCEIwoSP}{{\sf LocalCE-I w/o SP}}
\newcommand{\GlobalCE}{{\sf GlobalCE-B\&I}}

\newcommand{\GCN}{{\sf{GCN}}}

\newcommand{\clustergcn}{{\sf{Cluster-GCN}}}

\newcommand{\cora}{{\emph{Cora}}}

\newcommand{\pubmed}{{\emph{PubMed}}}
\newcommand{\bail}{{\emph{Bail}}}
\newcommand{\german}{{\emph{German}}}

\newcommand{\facebook}{{\emph{FacebookPagePage}}}
\newcommand{\amazon}{{\emph{AmazonProducts}}}

\newcommand{\fairgnn}{{\sf{FairGNN}}}
\newcommand{\hs}{{\sf{HS}}}

\newcommand{\sfield}{{\sf{SF}}}
\newcommand{\ia}{{\sf{IA}}}

\newtheorem{defn}{Definition}

\newtheorem{problem}{Problem}

\newcommand{\spara}[1]{\smallskip\noindent{\bf #1}}

\newcommand{\squishlist}{
 \begin{list}{$\bullet$}
  {  \setlength{\itemsep}{0pt}
     \setlength{\parsep}{3pt}
     \setlength{\topsep}{3pt}
     \setlength{\partopsep}{0pt}
     \setlength{\leftmargin}{2em}
     \setlength{\labelwidth}{1.5em}
     \setlength{\labelsep}{0.5em}
} }
\newcommand{\squishlisttight}{
 \begin{list}{$\bullet$}
  { \setlength{\itemsep}{0pt}
    \setlength{\parsep}{0pt}
    \setlength{\topsep}{0pt}
    \setlength{\partopsep}{0pt}
    \setlength{\leftmargin}{2em}
    \setlength{\labelwidth}{1.5em}
    \setlength{\labelsep}{0.5em}
} }

\newcommand{\squishdesc}{
 \begin{list}{}
  {  \setlength{\itemsep}{0pt}
     \setlength{\parsep}{3pt}
     \setlength{\topsep}{3pt}
     \setlength{\partopsep}{0pt}
     \setlength{\leftmargin}{1em}
     \setlength{\labelwidth}{1.5em}
     \setlength{\labelsep}{0.5em}
} }

\newcommand{\squishend}{
  \end{list}
}

\AtBeginDocument{%
  }

\acmConference[KDD '25]{Proceedings of the 31st ACM SIGKDD Conference on Knowledge Discovery and Data Mining V.2}{August 3--7, 2025}{Toronto, ON, Canada}
\acmBooktitle{Proceedings of the 31st ACM SIGKDD Conference on Knowledge Discovery and Data Mining V.2 (KDD '25), August 3--7, 2025, Toronto, ON, Canada}
\acmDOI{10.1145/3711896.3736960}
\acmISBN{979-8-4007-1454-2/2025/08}
\begin{document}

\title{Finding Counterfactual Evidences for Node Classification}

\author{Dazhuo Qiu}
\affiliation{%
  \institution{Aalborg University}
  \city{Aalborg}
  \country{Denmark}
}
\email{dazhuoq@cs.aau.dk}

\author{Jinwen Chen}
\affiliation{%
  \institution{University of Electronic Science and Technology of China}
  \city{Chengdu}
  \country{China}
}
\email{jinwenc@std.uestc.edu.cn}

\author{Arijit Khan}
\affiliation{%
  \institution{Aalborg University}
  \city{Aalborg}
  \country{Denmark}
}
\email{arijitk@cs.aau.dk}

\author{Yan Zhao}
\orcid{0000-0002-0242-3707}
\affiliation{%
  \institution{Shenzhen Institute for Advanced Study, University of Electronic Science and Technology of China}
  \city{Shenzhen}
  \country{China}
}
\email{zhaoyan@uestc.edu.cn}

\author{Francesco Bonchi}
\affiliation{%
  \institution{CENTAI}
  \city{Turin}
  \country{Italy}
}
\affiliation{%
  \institution{Eurecat}
  \city{Barcelona}
  \country{Spain}
}
\email{bonchi@centai.eu}


\begin{abstract}
Counterfactual learning is emerging as an important paradigm, rooted in causality,
which promises to alleviate common issues of graph neural networks (\GNNs), such as fairness and interpretability. However, as in many real-world application domains where conducting randomized controlled trials is impractical, one has to rely on available
observational (factual) data to detect counterfactuals.
In this paper, we introduce and tackle the problem of searching for
counterfactual evidences for the \GNN-based node classification task.
A counterfactual evidence is a pair of nodes such that, regardless they exhibit great similarity both in the features and in their neighborhood subgraph structures, they are classified differently by the \GNN. We develop effective and efficient search algorithms and a novel indexing solution that leverages both node features and structural information to identify counterfactual evidences, and generalizes beyond any specific \GNN. Through various downstream applications, we demonstrate the potential of counterfactual evidences to enhance fairness and accuracy of \GNNs.
\end{abstract}



\received{20 February 2025}
\received[revised]{18 April 2025}
\received[accepted]{16 May 2025}

\maketitle

\section{Introduction}\label{sec:introduction}

Graph neural networks (\GNNs) have demonstrated significant performance and representational capacity in solving problems over graph-structured data, fueling a great deal of applications in a variety of domains, ranging from biology to finance, from social media to neuroscience \cite{wu2022graph}. In particular, among various graph machine learning tasks, \GNNs\ have shown significant predominance over other methods in the \emph{node classification} task, that is, when a unique graph is given as input with some nodes labelled and some are not, and the goal is to predict a label for each unlabeled node.

Together with the demonstrated predictive performance, \GNNs\ inherit from the wider class of deep learning methods several limitations which can hinder their adoption, especially in high-stakes application domains \cite{DaiZZXGLTW24}.
Most noticeably, \GNNs\ lack interpretability \cite{DaiW21b,YuanYGJ23} and tend to inherit biases present in the training datasets \cite{DaiW21,KoseS22,DongLJL22,WangZDCLD22,DaiW23,KoseS24,KoseS24b}, potentially leading to discriminatory decisions based on, e.g., gender, race, or other sensitive attributes.
The latter issue derives from the fact that, exactly like every other machine learning algorithm, \GNNs\ are trained to reflect the distribution of the training data which often contains historical bias towards sensitive attributes. If not addressed explicitly, the bias contained in the training data can end up being structured in the trained machine learning model \cite{HajianBC16}. Furthermore, the underlying graph structure and the typical message-passing mechanism of \GNNs\ can further magnify harmful biases \cite{DaiW23}.

\emph{Counterfactual learning} is emerging as a paradigm that can alleviate both fairness and interpretability issues \cite{guo2023counterfactual}. The notion of counterfactual is borrowed from causal language and it indicates the possibility of an alternative outcome if some of the premises were different from what were in reality  (``counter to the facts''). For instance, in fairness testing, counterfactual examples are created by altering certain sensitive features (e.g., gender or race) of a data point to see if the model's predictions change. Counterfactuals are also straightforward contrastive example-based explanations of the type:
\emph{``You were denied a loan. If your income had been \pounds 45,000, you would have been offered a loan''}~\cite{counterfactuals}.
A key challenge for learning causality \cite{GuoCLH020} is that, in order to properly determine the causal effect of an action, we need to know both the factual outcome with the observed action and the counterfactual outcome with the unobserved action. However, in many real-world settings it is impractical to conduct randomized controlled trials to get the counterfactual: When this is the case, we only have access to
the observational factual data, i.e., the observed
action and its corresponding factual outcome. Therefore, \emph{it is crucial to develop methods for detecting counterfactuals in the observational data}, in order to take the full advantage of counterfactual reasoning in machine learning \cite{pearl2016causal,GuoCLH020,abs-2206-15475,guo2023counterfactual}.

In this paper, we tackle the problem of \emph{searching counterfactual evidences in node classification task}. In this setting, a counterfactual evidence for a node $v$ is another node $u$ such that, regardless they exhibit great similarity in their neighborhood subgraph, including the features, they are classified differently by the given \GNN. The differences between $v$'s and $u$'s neighborhood subgraphs, are what define the counterfactual hypothesis for $v$, i.e., the small changes that would make $v$ being classified differently. The advantage of this type of counterfactual over perturbation-based ones, is that \emph{it exists in the factual data}: as such it enjoys greater realism and feasibility than counterfactuals produced by perturbation.

\begin{figure}[t!]
    \centering
    \Description[intro example]{}{}
    \includegraphics[width=0.96\linewidth]{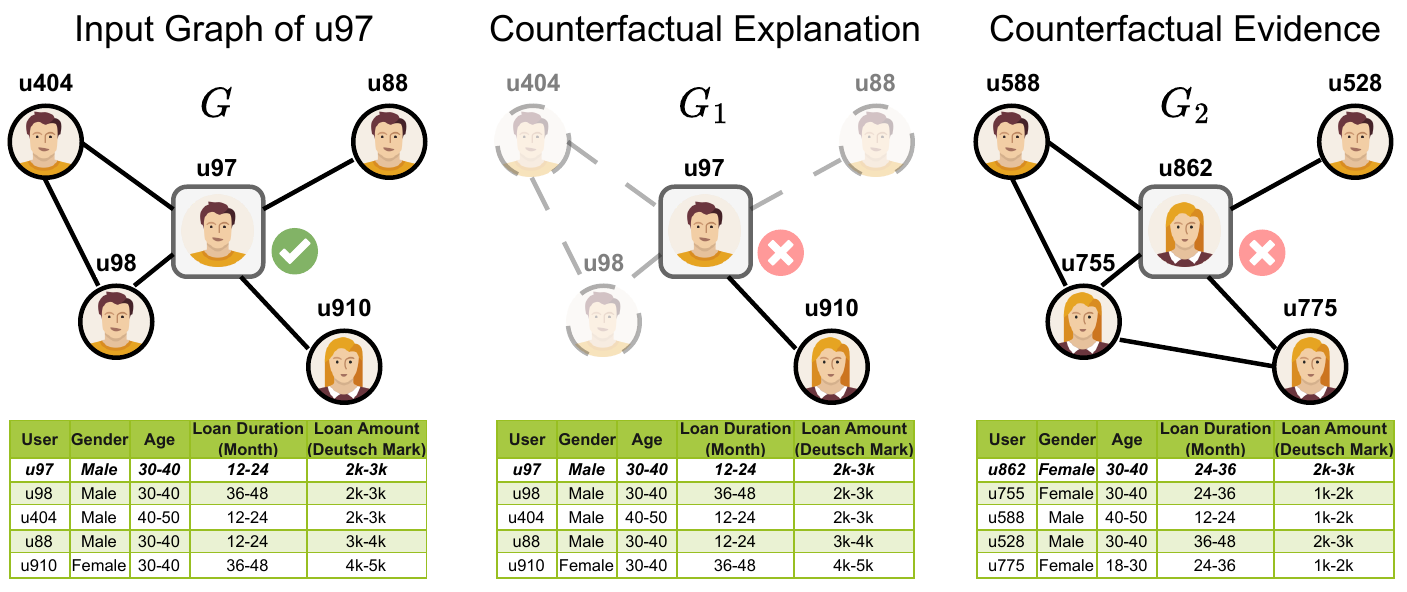}
    \caption{\small
    (Left) The 1-hop neighborhood $G$ of the node of interest, $u97$, from {\em German Credit} dataset~\cite{nifty}.
    Each customer has important features for classification, as shown in the table.
    (Center) $G_1$ is a counterfactual explanation induced from $G$, highlighting the important parts of the input graph.
    (Right) $u862$ is the counterfactual evidence that shares similar node features and surrounding structures ($G_2$) with $u97$, but has a different classification result.}
    \label{fig:intro_example}
\end{figure}

Figure~\ref{fig:intro_example} provides a depiction of a counterfactual evidence and its difference from a \emph{counterfactual explanation}~\cite{YuanYGJ23}, extracted from the {\em German Credit} dataset~\cite{nifty}. On the left, a node, $u97$, is presented together with its 1-hop neighborhood structures ($G$), along with feature values for all the nodes in $G$. In this case, the \GNN\ classifies $u97$ positively (loan approved).
In the center, a counterfactual explanation for $u97$ is presented: this is a set of perturbations to the specific data point ($u97$'s 1-hop neighborhood
structure and features), which is sufficient to induce a change in the classification.
In this example, the counterfactual explanation (subgraph $G_1$) is obtained from the data point $G$ by masking three out of four links: such changes would make $u97$ classified negatively, thus highlighting which are the most important parts of $G$ for the prediction.
Instead, a counterfactual evidence is a different node ($u862$ on the right), which regardless of having great similarity in the subgraph and associated features, gets a different treatment from $u97$.

Besides the general importance of finding counterfactuals in the observational data (previously discussed), there are some immediate applications of graph counterfactual evidence that we develop in case studies in \S \ref{sec:applications}.
From the standpoint of a watchdog auditing an algorithmic decision-making system for unlawful discrimination, it is critical to find evidences of nodes which are similar, but treated differently, thus violating the principle that “similar individuals should be treated similarly”~\cite{DworkHPRZ12}, indicating \emph{potential unfairness issues} with the model.
Two similar yet classified differently nodes might also be an indication of an area around the decision boundary of the \GNN, where misclassification errors concentrate, thus providing a signal for potential intervention to \emph{improve the performance of a classifier}.

\spara{Our contributions.} We tackle the problem of extracting, from a given graph, a pair of nodes (together with their neighborhoods), such that they exhibit high similarity in both structure and features, yet they are classified differently by a \GNN.
Our is a data mining problem, that is, we define a novel structure of interest in a large graph based on a pre-trained GNN model's output and we then devise the algorithm to extract it.
More specifically, we consider two versions of the problem: the \emph{local} version seeks counterfactual evidences for a given target node, while the \emph{global} version seeks for counterfactual evidences among all possible node pairs. For both problems, we aim to extract the top-$k$ counterfactual evidences w.r.t. a measure of similarity of the two neighborhood subgraphs. As a measure of similarity, we adapt the Weisfeiler Lehman (\WL) kernel-based graph similarity \cite{wltest, wlkernel}, for the case of node-anchored graphs (i.e., $L$-hop neighbor subgraphs), having multiple node features.  Recall that the \WL-test is a necessary but insufficient condition for graph isomorphism~\cite{wltest},
whereas the recent \GNN\ variants are not more powerful than the 1-\WL\;test~\cite{gin}. Our technique inherits the simplicity and computational efficiency of the \WL\;kernel computation, while following the update scheme of message-passing \GNNs, thus it generalizes the inference process without being tailored for any specific \GNN.

We propose search algorithms for both local and global problems, which are similar to exact and approximate nearest neighbor search over high-dimensional, dense vector spaces \cite{index_ref2}. First, we design a baseline algorithm that applies a linear scan over all test nodes, and we utilize several optimization strategies.
Second, we propose an index-based algorithm to improve the efficiency by pruning undesirable test nodes.
Existing vector indexes ~\cite{lsh, hnsw} are suitable for Euclidean distance, they are not readily applicable to cosine similarity that we adopt, thereby requiring a novel, efficient, and effective index tailored for our problem.
Specifically, our novel index-based algorithm creates {\em supplementary clusters} based on new centroids that are close to the boundary nodes, thereby enhancing the quality of index-based retrieval at the cost of some redundancy.

Our experiments assess the efficiency and effectiveness of our algorithms on real-world datasets, using different \GNNs\, and comparing against non-trivial baselines. Finally, we showcase the potential of applying counterfactual evidences in various downstream tasks: (1) we show how our proposal can be effective in unveiling unfairness of a \GNN; (2)  we show how counterfactual evidences identify the test instances that are close to the decision boundary of a \GNN\ and thus error-prone; (3)
 we illustrate that fine-tuning the \GNN\ with counterfactual evidences can enhance  
 accuracy.

\spara{Paper contributions and roadmap:} 
\squishlist
\item We introduce the novel notion of counterfactual evidence for node classification (\S\ref{sec:preliminaries}).

\item We propose a novel and generic kernel-based graph similarity measure to assess the similarity between neighborhood subgraphs for a pair of nodes (\S\ref{sec:ks}).

\item We introduce an index-based algorithm to efficiently search counterfactual evidences while maintaining high quality (\S\ref{sec:method}).

\item We assess our algorithms on several real-world datasets, using different \GNNs\, and against non-trivial baselines (\S\ref{sec:exp}). Finally, we showcase the potential of applying counterfactual evidences in various downstream tasks (\S\ref{sec:applications}).

\squishend

\section{Related work} 

In graph machine learning, the term ``counterfactual'' is adopted in different contexts,  where it takes different semantics \cite{guo2023counterfactual,PradoRomeroPSG24}. We next review some related literature highlighting how the notions of counterfactual differ from the novel notion of counterfactual evidence in node classification, that we introduce in this paper.

\spara{Fairness.} The notion of 
\emph{counterfactual fairness} is based on the idea that a prediction for an instance
is fair if it remains the same in a counterfactual world where the instance belongs to a different protected group, e.g., a different gender or race. Counterfactual learning on graphs has emerged as a promising direction to achieve counterfactual fairness, which is attained via a trade-off between utility and fairness in the objective function \cite{nifty,gear,ZhangZJL21}. Our focus is different: We retrieve counterfactual nodes as opposed to achieving counterfactual fairness in \GNN\ classification. Nevertheless, we demonstrate in our empirical evaluation that our counterfactual evidences can facilitate detecting unfairness patterns in \GNNs\ (\S\ref{sec:applications}).

\spara{Explainability.} For \GNN\ explainability, a {\em factual explanation} is a subgraph that preserves the result of classification, while a {\em counterfactual explanation} is a subgraph which flips the result if perturbed or removed \cite{YuanYGJ23}.
A number of counterfactual explainability methods for \GNNs\ have been proposed considering both node and graph classifications, e.g., \textsf{CF-GNNExplainer} \cite{LucicHTRS22}, \textsf{RCExplainer} \cite{BajajCXPWLZ21}, \textsf{MEG} \cite{numeroso2021}, and \textsf{CLEAR} \cite{MaGMZL22}. Recent works combine factual and counterfactual explanations for \GNNs\ \cite{TanGFGX0Z22,ChenQWKKG24,QiuW0W24}.
These methods identify a subgraph surrounding a node such that, if the subgraph is perturbed, the node classification will be different. This notion of counterfactual thus differs from the counterfactual evidence that we propose in this paper, as already discussed in \S 1 and depicted in Figure 1.

In the context of \emph{graph classification}, Abrate et al. introduce the notion of {\em counterfactual graph}, which has high structural similarity with the input graph, but is classified by the \GNN\ into a different class \cite{AB21}. Huang et al. propose \textsf{GCFExplainer} to identify a small set of representative counterfactual graphs, referred to as the {\em global counterfactuals}, for all input graphs in a graph database \cite{HKMRS23}. Both these works share with our search paradigm for counterfactuals in the observational data, but they focus on graph classification (instead of node classification, as we do in this paper): they consider graph structures and use symmetric difference or graph edit distance to measure similarity between graph pairs, while they disregard multiple node features, which instead play an important role in our proposal.

\section{Problem statement}
\label{sec:preliminaries}
%
%
We consider an undirected unweighted \textbf{graph}  $G=(V, E)$, where $V=\{v_1,v_2,...,v_{|V|}\}$ denotes a finite set of nodes and $E \subseteq V \times V$ a set of edges. Nodes are associated with $d$-dimensional features, represented by the feature matrix $\mathbf{X} \in \mathbb{R}^{|V|\times d}$, where $\mathbf{x}_v\in \mathbb{R}^d$ is the $d$-dimensional feature vector associated with a node $v \in V$. Next, $\mathbf{Y}$ denotes a set of true (ground truth) labels associated with nodes, and $y_v \in \mathbf{Y}$ is the true label for node $v$.
\\
\textbf{Graph Neural Networks ({\sf GNNs})}~\cite{gcn,du2021multi,jiang2025ICML} comprise a well-established family of deep learning models tailored for analyzing graph-structured data. {\sf GNNs} generally employ a multi-layer message-passing scheme as shown in Equation~\ref{eq-gnn1}.
\begin{equation}
\small
\label{eq-gnn1}
    \mathbf{H}^{(l+1)} = \sigma(\widetilde{\mathbf{A}}\mathbf{H}^{(l)}\mathbf{W}^{(l)})
\end{equation}
$\mathbf{H}^{(l+1)}$ is the matrix of node representations at layer $l$, with $\mathbf{H}^{(0)}=\mathbf{X}$ being the input feature matrix. $\widetilde{\mathbf{A}}$ is the normalized adjacency matrix of the graph, which captures the graph structure. $\mathbf{W}^{(l)}$ is a learnable weight matrix at layer $l$. $\sigma$ is an activation function such as {\sf ReLU}. The final layer's output $\mathbf{H}^{(L)}$ is used to make predictions by passing it to fully-connected and then {\sf softmax} layers.

\textbf{Node classification} is a fundamental task in graph analysis~\cite{nc}. In the context of {\sf GNNs}, it aims to learn a model $M: V \rightarrow \mathbf{Y}$ s.t. $M(v)=y_v$ for $v \in V_{train} \subseteq V$, where $V_{train}$ is the training set of nodes with known (true) labels $\mathbf{Y}_{train}$. Then, we tune the hyper-parameters of the model using a validation set of nodes $V_{valid}$ and their known labels $\mathbf{Y}_{valid}$. Finally, the \GNN\ predicts the labels for the remaining test nodes $V_{test}=V\backslash (V_{train} \cup V_{valid})$, also known as the inference mechanism of \GNNs.

We consider a fixed, deterministic \GNN\ $M$, that is, (1) it has all factors which determine the inference process of $M(\cdot)$ such as layers, model parameters, etc. fixed; and (2) $M(\cdot)$ always generates the same output label for the same input test node.

The \textbf{$L$-hop neighbor subgraph} of a node $v \in V$ is a subgraph $G_v \subseteq G$, where each node in $G_v$ can be reached from $v$ within $L$-hops. This subgraph is anchored by node $v$. In a message-passing \GNN\;with $L$-layers, each layer aggregates messages from neighboring nodes. With each successive layer, the messages are propagated one step further. Thus, after $L$ layers, a node $v$ can aggregate information from its $L$-hop neighbors, emphasizing the importance of the $L$-hop neighbor subgraph $G_v$ surrounding the node $v$.

\textbf{Counterfactual Evidence.}
    Given a node $v \in V_{test}$,
    another node $u \in V_{test}, (v\neq u)$ is called its counterfactual evidence ({\CE}) if the following two conditions hold:
    \\
    1) $v$ and $u$ are assigned different labels by $M$: i.e., 
       $M(v)\neq M(u)$;
  \\  
    2) The $L$-hop neighbor subgraphs of $v$ and $u$ have a high similarity score (``$\KS(v, u)$ measure'' that we define in the next section).
    
Given that we search for pairs with high similarity, one option could be to retrieve all the pairs having similarity score above a given threshold. In this paper instead, we seek for the top-$k$ pairs. For the sake of simplicity of presentation, we formalize both local and global versions of the problem for top-$1$, next.
    
\begin{problem}[Top-$1$ Local Counterfactual Evidence]
Given a query node $v \in V_{test}$, the top-$1$ counterfactual evidence, $\LCE_{opt}(v)$ is a node $u \in V_{test}$ that 1) has a different predicted label w.r.t. $v$; and 2) attains the highest similarity score $\KS(v, u)$ compared to all other nodes in the test set.
\begin{equation}
\small
\label{top1}
   \LCE_{opt}(v) = \argmax_{u \in V_{test}, \, \, M(v) \ne M(u)} \KS(v, u)
\end{equation}
\end{problem}
\begin{problem}[Top-$1$ Global Counterfactual Evidence]
Given the test set $V_{test}$, the top-$1$ global counterfactual evidence, $\GCE_{opt}(V_{test})$ is a pair of nodes $(v, u), (v,u \in V_{test}$) s.t. 1) $v$ and $u$ have different predicted labels; and 2) the pair has the highest similarity score $\KS(v, u)$ among all possible node pairs in $V_{test}$.
\begin{equation}
\small
\label{gtop1}
   \GCE_{opt}(V_{test}) = \argmax_{v, u \in V_{test}, \, \, M(v) \ne M(u)} \KS(v, u)
\end{equation}
\end{problem}
\noindent
The generalization of \LCE\ and \GCE\ problems to extracting the {\em top-$k$} pairs is straightforward. 
\section{Kernel-based Similarity}
\label{sec:ks}
The similarity between two graphs can be computed in many different ways~\cite{sim_dif}. Purely structural methods, e.g., graph isomorphism~\cite{si}, graph edit distance~\cite{GEDsurvey}, and maximum common subgraph~\cite{mcs1, mcs2} are relatively strict and may ignore node features. GNN-based approaches such as NTN~\cite{gnn_sim2}, SimGNN~\cite{simgnn}, and MGMN~\cite{gnn_sim3} would be specific to the underlying \GNN s, and require supervision including extensive parameter tuning and training epochs. In this work, we instead adapt the Weisfeiler Lehman (\WL) kernel-based graph similarity measure \cite{wltest, wlkernel}, for the case of node-anchored graphs (i.e., $L$-hop neighbor subgraphs), having multiple node features.  Recall that the \WL-test is a necessary but insufficient condition for graph isomorphism~\cite{wltest},
whereas the recent \GNN\ variants are not more powerful than the 1-\WL\;test~\cite{gin}. We refer to our technique as the {\em kernel-based similarity} (or, \KS, in short).  It inherits the simplicity and computational efficiency of the \WL\;kernel computation. Meanwhile, our approach \KS\ also follows the update scheme of message-passing \GNN s, thus it generalizes the inference process of \GNN s, without being tailored for a specific \GNN.

To this end, we first introduce the traditional \WL\;kernel-based graph similarity measure \cite{wltest, wlkernel}.
In this setting, unlike multiple features as in \GNN-based approaches (\S\ref{sec:preliminaries}), each node $v$ has one single attribute $a_v$. The \WL\ kernel identifies similarities by examining subtree patterns. These patterns emerge from a propagation scheme that iteratively evaluates the attributes of nodes and their neighbors. The process involves generating a sequence of ordered strings by aggregating the attributes of a node with those of its neighbors, which are subsequently hashed to generate new, compressed node attributes.
In each iteration, these updated attributes represent expanded neighborhoods of each node, e.g., at iteration $L$, the compressed label summarizes $L$-hop neighborhoods.
Specifically, the initial attribute $a_v$ of each node $v$ refers to $a_v^0$,
and let $L$ denote the number of {\sf WL} iterations.
For the $l$-th iteration,
we update the compressed attribute $a_v^l$ by looking at the ordered set of neighbor attributes, as depicted in Equation~\ref{wlscheme}.
\begin{equation}
\small
    a_v^l = hash(a_v^{l-1}, N^{l-1}(v))
    \label{wlscheme}
\end{equation}
$N^l(v)$ denotes the sorted sequence of attributes at the $l$-th iteration from the 1-hop neighbors of node $v$. The hash function generates an updated, compressed attribute for node $v$. For this purpose, perfect hashing is utilized, ensuring that two nodes at the $l$-th iteration possess identical attributes if and only if their attributes and those of their neighbors at the $l$-th iteration are identical.

Our problem setting is different from the classic \WL\ kernel in two ways. First, we compare the similarity between node-anchored graphs (\S\ref{sec:preliminaries}), which effectively preserves the information of both the target nodes and their neighborhoods. Second, instead of focusing on a single attribute as in the \WL\ kernel, we incorporate multiple node features to better capture the characteristics of the nodes and their neighbors. This approach allows us to represent a node with both rich information and neighborhood dependencies. Therefore, we define the \KS\ score as below.
\begin{defn}[\KS\ score]
We incorporate information about multiple node features in the classic {\sf WL} scheme, that is, unlike updating node colors in the original {\sf WL} kernel, we update the node features of each node based on its current vector and the vectors of its neighbors.
Specifically, we compute a weighted sum of the neighbor vectors using the cosine similarity as weights, then combine it with the original feature vector based on a trade-off parameter $\alpha$, as follows.
    \begin{equation}
    \small
        \mathbf{x}^{l+1}_v = \alpha \cdot \mathbf{x}^{l}_v + \frac{1-\alpha}{|N(v)|} \sum_{u \in N(v)} {\sf COSINE}(\mathbf{x}^{l}_v, \mathbf{x}^{l}_u) \cdot \mathbf{x}^{l}_u
        \label{ks_update}
    \end{equation}
    \begin{equation}
    \small
       {\sf COSINE}(\mathbf{x}^{l}_v, \mathbf{x}^{l}_u) = \frac{\mathbf{x}^{l}_v \cdot \mathbf{x}^{l}_u}{||\mathbf{x}^{l}_v||_2 \cdot  ||\mathbf{x}^{l}_u||_2}
       \label{csk}
   \end{equation}
    Here, $N(v)$ represents the 1-hop neighbors of node $v$, $\mathbf{x}^{l}_v$ and $\mathbf{x}^{l}_u$ are the feature vectors of node $v$ and node $u$ at iteration $l$, respectively, with $\mathbf{x}^{0}_v=\mathbf{x}_v$. The trade-off parameter $\alpha$ controls the relative importance between a node's original feature vector and its accumulated (weighted) feature vectors from the neighbors.
    \\
    To capture the evolution of the graph structure across iterations, we aggregate the vectors across multiple iterations for a node, that is, $\mathbf{x}^{agg}_v = \sum_{l=0}^{L}\mathbf{x}^{l}_v$.
    Recall that $L$ is obtained from the $L$-hop neighbor subgraph surrounding each node. Finally, we define the kernel-based similarity $\KS(v,u)$ as the cosine similarity between $\mathbf{x}^{agg}_v$ and $\mathbf{x}^{agg}_u$.
    \begin{equation}
    \small
        {\sf KS}(v, u) = \frac{\mathbf{x}^{agg}_v\cdot \mathbf{x}^{agg}_u}{||\mathbf{x}^{agg}_v||_2\cdot ||\mathbf{x}^{agg}_u||_2}
    \end{equation}
\label{ks_sim}
\end{defn}
\noindent
We employ the cosine similarity, since compared to the Euclidean distance between two vectors, cosine similarity is invariant to the magnitude of the vectors and also ranges between $(0,1)$ -- which simplifies the process for end users.

\section{Algorithms}
\label{sec:method}
In this section, we present our proposed algorithms for identifying counterfactual evidences. We begin by outlining the baseline algorithm for local \CE\ identification, including its optimization strategies. Next, we develop an index-based algorithm designed to enhance the efficiency of \CE\ identification. Finally, we extend the algorithm for discovering global counterfactuals. 
\subsection{Local \CE\ Identification}
\label{sec:localCEmethod}
\begin{algorithm}[tb!]    
\small
\renewcommand{\algorithmicrequire}{\textbf{Input:}}
    \renewcommand{\algorithmicensure}{\textbf{Output:}}
    \caption{\LocalCEB}
    \begin{algorithmic}[1]
        \REQUIRE Graph $G$, \GNN\ $M$, test set $V_{test}$, query node $v$.
        \ENSURE Top-$1$ local counterfactual evidence $\LCE_{opt}(v)$ for $v$.  
        \STATE $\LCE_{opt}(v) := \texttt{None}$.
        \FOR{$u \in V_{test}\backslash v$}\label{linear}
            \IF{$M(v)\neq M(u)$ and $\KS(v, u)>\KS(v, \LCE_{opt}(v))$\label{criteria1}}
            
                \STATE $\LCE_{opt}(v) := u$\label{optimal}.
            \ENDIF
        \ENDFOR
        \RETURN $\LCE_{opt}(v)$\label{return1}. 
    \end{algorithmic}
  \label{alg:LocalCEB}
\end{algorithm}
\textbf{Baseline.}
We first propose the baseline algorithm for finding the top-$1$ local counterfactual evidence, denoted as \LocalCEB. As shown in Algorithm~\ref{alg:LocalCEB}, given a graph $G$, a \GNN\ model $M$, a subset of test nodes $V_{test}$ for classification, and a query node $v$, \LocalCEB\ performs a linear scan over the remaining nodes (line~\ref{linear}) to identify the top-$1$ $\LCE_{opt}$ that satisfies the specified criteria (line~\ref{criteria1}). A node $u$ is considered $\LCE_{opt}$ if: 1) the predicted labels of $u$ and $v$ by the pre-trained {\sf GNN} model $M$ are different; and 2) the \KS\ score between $v$ and $u$ is the largest. \LocalCEB\ iteratively updates $\LCE_{opt}(v)$ with the node that achieves the highest \KS\ score (line~\ref{optimal}). Finally, \LocalCEB\ outputs the top-$1$ local counterfactual evidence.

Extending the algorithm to find the top-$k$ \LCE\ is straightforward. We maintain a bucket $B$ of size $k$, where nodes are sorted in descending order based on their \KS\ scores. If 
a new incoming node has a \KS\ score higher than the last node in $B$, we remove the last node and insert the new node into the bucket, keeping the nodes sorted in descending order.

\underline{\emph{Time and Space Complexity.}}
The time complexity of \LocalCEB\ consists of three main parts, first is the inference cost of \GNN: 
$O(Ld(|E|+|V_{test}||\mathbf{Y}|))$ 
\cite{complex_survey} assuming \GCN~\cite{gcn}, \clustergcn~\cite{clusterGCN}, etc.; second is the cost for the vector aggregation of \KS\ score: $O(Ld|V_{test}||N(\cdot)|)$ with $|N(\cdot)|$ being the maximum number of 1-hop neighbors of a test node; finally is the cost of finding the top-$k$: $O(k|V_{test}|)$. Therefore, the total time complexity of \LocalCEB\ is $O\left(Ld\left(|E|+|V_{test}|\left(|\mathbf{Y}|+\right.\right.\right.$ $\left.\left.\left.|N(\cdot)|\right)\right)+k|V_{test}|\right)$. As for the space complexity, the space cost of \GNN's inference is $O\left(d|V_{test}|+|E|\right)$ \cite{complex_survey}; the space cost of vector aggregation is $O(d|N(\cdot)||V_{test}|)$, and for top-$k$ is $O(k)$. Therefore, the total space complexity of \LocalCEB\ is $O(d|N(\cdot)||V_{test}|+|E|+k)$.

\underline{\emph{Optimizations.}}
We apply two optimization strategies for the \LocalCEB\ algorithm: {\bf (1)} The prediction of each test node is pre-computed by the inference of \GNN. Additionally, the test nodes are partitioned based on their predicted classes and stored separately, thus at the query time we only look at the nodes from different classes. {\bf (2)} We also pre-compute the aggregated vectors of all test nodes, so we can rapidly identify the top-$1$ \LCE\ by a linear scan over the aggregated vectors of test nodes from different classes. 

\textbf{Index-based Solution.}
Performing a linear scan over aggregated vectors of test nodes to compute cosine similarity is computationally expensive, particularly when dealing with large-scale graphs. This inefficiency poses a significant challenge in applications such as information retrieval and recommendation systems, where rapid and accurate similarity calculations are crucial. Existing vector index approaches~\cite{lsh, hnsw}, which are primarily optimized for Euclidean distance, are not readily applicable to cosine similarity, further complicating the problem. Additionally, there are almost no dedicated index approaches specifically designed for cosine similarity~\cite{index_ref1,index_ref2}. This gap necessitates the development of an efficient and effective solution tailored for cosine similarity computations. To address these challenges, we propose a novel heuristic index-based algorithm that leverages the $k$-means clustering and is designed for cosine similarity. 
Specifically, we make two novel technical contributions: {\bf (1)} \emph{Supplementary partitioning}
that enhances the quality of our index, and {\bf (2)} \emph{weighted clustering} to adapt the $k$-means algorithm to generate supplementary partitions.  Based on these novel techniques, we develop an index structure and show how querying benefits using this structure.

\underline{\emph{Overview.}}
Our core idea is to apply the $k$-means algorithm that assigns vectors to different clusters. For a specific query vector, we identify its cluster and find the top-$k$ \CEs\ only from that cluster, reducing the search space by a factor of the number of clusters.

However, a high-quality result requires the query vector to be near the cluster centroid, since its top-$k$ \CEs\ are then expected to be in the same cluster. In contrast, if the query vector is near the cluster boundary, some top-$k$ \CEs\ might belong to a different cluster, leading to higher errors, as we will miss those \CEs.

To address this limitation, we propose \emph{supplementary partitioning}. This method creates more partitions where boundary nodes
from earlier partitions are assigned closer to some centroid in the new partition. For a query, we first select the optimal partition, and
then find its cluster according to that partition, finally identify the top-$k$ \CEs\ from that cluster, thereby improving the quality of results despite some storage redundancy due to multiple partitioning.

We construct supplementary partitions using weighted k-means clustering, assigning more weight to nodes closer to their cluster boundaries from the previous partitions. These weights make boundary nodes more central in the new partition.

\underline{\emph{Supplementary Partitioning.}} 
Consider a previously computed partition $C^0=\{\mathbf{c}^0_1, \mathbf{c}^0_2, \ldots, \mathbf{c}^0_m\}$, where $m$ is 
the number of clusters. Based on this partitioning, a weight $w^0_v$ is assigned to each test node $v$. For simplicity, assume that $w^0_v$ is proportional to the distance of $v$ from the centroid of $v$'s assigned cluster. More details on the weight assignment is given later.

Next, we adapt the $k$-means algorithm to incorporate the weight
of each node so that the previous boundary nodes are assigned closer to some centroid in the new 
partition $C^1=\{\mathbf{c}^1_1, \mathbf{c}^1_2, \ldots, \mathbf{c}^1_m\}$. Thus, the objective function of the classic $k$-means is updated.
\begin{eqnarray}
\small
    \arg\max_{C^1} \sum_{i=1}^m\sum_{v\in\mathbf{c^1_i}}w^0_v \cdot {\sf COSINE}(\mathbf{x}^{agg}_v,\mu^1_i)\\
   \mu^1_i = \frac{1}{|\mathbf{c^1_i}|} \sum_{v\in\mathbf{c^1_i}} w^0_v \cdot\mathbf{x}^{agg}_v
    \label{wkmeans}
\end{eqnarray}

$\mathbf{x}^{agg}_v$ is the aggregated vector of node $v$. 
Based on such weighted $k$-means, we obtain the new partition $C^1$, one of its newly computed centroids $\{\mu^1_1, \mu^1_2, \ldots, \mu^1_m\}$ will have more chance to be closer to the previous boundary nodes from partition $C^0$. 
\begin{figure}[t!]
    \centering
    \Description[circle]{}{}
    \includegraphics[width=0.96\linewidth]{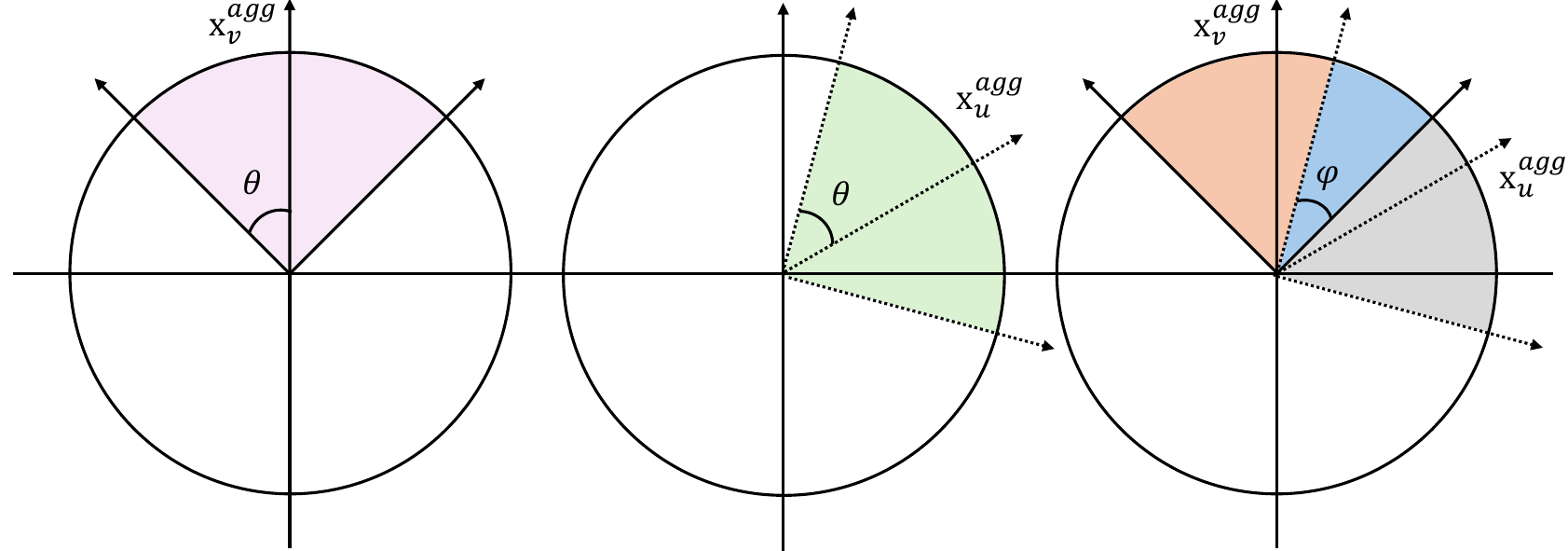}
    \vspace{-2mm}
    \caption{\small Example of vector intersection in $2$-dimension}
    \label{fig:circle}
    \vspace{-2mm}
\end{figure}

\underline{\emph{Weighted Clustering.}} 
To compute node weights discussed in the previous part, we propose a novel weight computation approach that incorporates the geometric property of the corresponding aggregated vectors. For simplicity, assume that the vectors are unit vectors since the cosine similarity computation already considers unit vectors. We introduce the idea through an example with $2$-dimensional vectors in Figure~\ref{fig:circle}. 
First, for the aggregated vector of each test node, we set an angle $\theta$ surrounding it to identify similar vectors. 
As we can see from the first circle, constrained by $\theta$, all similar vectors of $\mathbf{x}^{agg}_v$ fall within the purple area. We refer to this area the \emph{Similar Field} (\sfield). Similarly, for $\mathbf{x}^{agg}_u$, the green area is the \sfield.
These two vectors are similar to each other since they fall into each other's \sfield, which is the blue area in the third circle. Notably, the size of the intersection area is determined by the angle $\phi$.
Meanwhile, the relative complement area (orange area in the third circle) of $\mathbf{x}^{agg}_v$ w.r.t. $\mathbf{x}^{agg}_u$ is the area where the most similar vector to $\mathbf{x}^{agg}_v$ might be found that are not most similar to $\mathbf{x}^{agg}_u$. 

Based on this observation, if we consider $\mu$ as a cluster centroid and $v$ as a node assigned to this cluster, the intersection area $\ia(\mu, \mathbf{x}^{agg}_v)$ between $\mu$ and $\mathbf{x}^{agg}_v$ represents the area where the most similar vector to $\mathbf{x}^{agg}_v$ is likely to be found within $v$'s assigned cluster. If we assume that the threshold angle $\theta$ is the same across all vectors, we can normalize the intersection area $\ia(\mu, \mathbf{x}^{agg}_v)$, by dividing it via the \sfield\ of $\mathbf{x}^{agg}_v$. This normalized intersection value is proportional to the probability that the most similar vector of $\mathbf{x}^{agg}_v$ falls within $v$'s assigned cluster. We utilize this ratio to compute a node weight $w_v$ for the subsequent supplementary partitioning. 
\begin{equation}
\small
    w_v = 1-\frac{\ia(\mu, \mathbf{x}^{agg}_v)}{\sfield(\mathbf{x}^{agg}_v)}
    \label{eq:hyperscore}
\end{equation}
Intuitively, $w_v$ captures the potential error when identifying the most similar vector of $\mathbf{x}^{agg}_v$ only from $v$'s assigned cluster.
In practice, the value of \sfield\ can be pre-computed, since the radius $r$ and threshold $\theta$ are constants. Meanwhile, the computation of $\ia(\mu, \mathbf{x}^{agg}_v)$ can be optimized by normalizing the vectors to unit vectors. The above equation can be extended to $d$-dimensional hyperspherical space, where $d$ is our embedding dimensionality. 
The formulation for computing the intersection in $d$-dimensional hyperspherical space is detailed in~\cite{hyper_inter,hyper_total}.
\begin{figure}[t!]
    \centering
    \Description[circle]{}{}
    \includegraphics[width=0.96\linewidth]{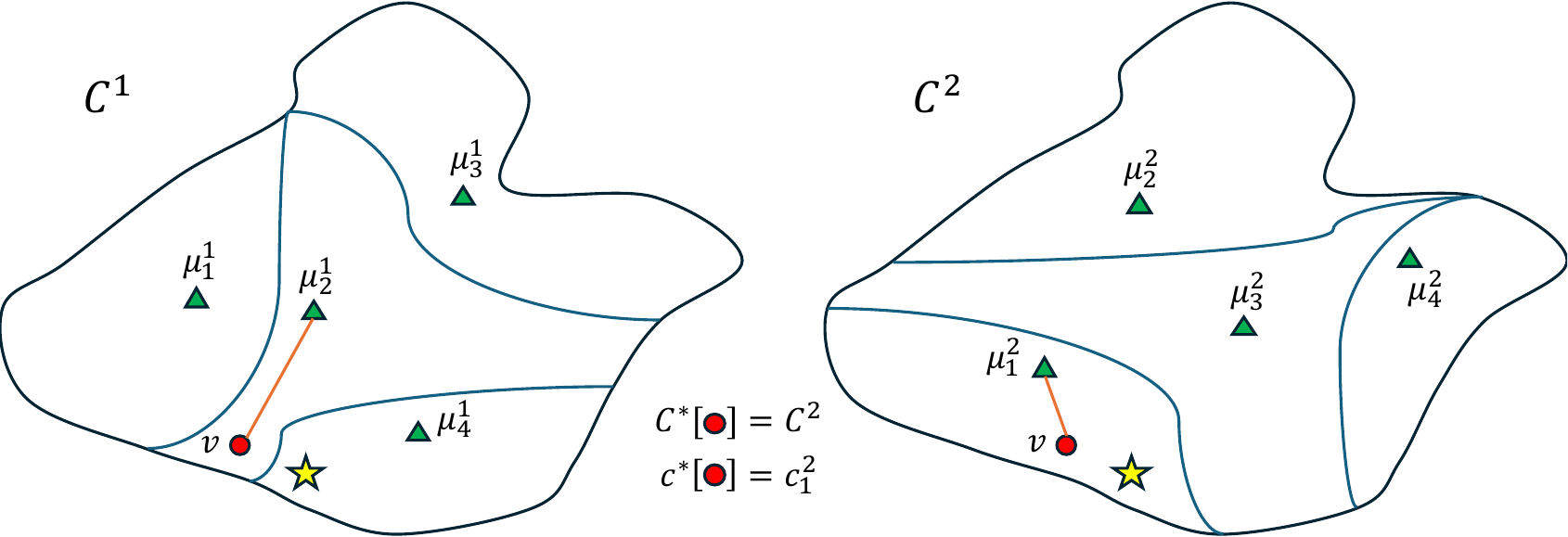}
    \caption{\small Supplementary partitioning: Two partitions $C^1$ and $C^2$, number of clusters in each partition $m=4$, green triangles are centroids of each cluster, the red node is the query node, the orange lines indicate the distance to the centroid and also reflect the weights, and the yellow star indicates the top-$1$ \LCE\ of the query node.}
    \label{fig:sp}
\end{figure}

\underline{\emph{Index Construction.}}
After the computation of partitions and clusters, assuming that we obtain total $p$ partitions, denoted as $P=\{C^1, C^2, \ldots, C^p\}$, we next aim to identify the optimal partition $C^*[v]$ and the optimal cluster $c^*[v]$ for each test node $v$. $C^*[v]$ is identified by the weight for each node $v$ at each partition from Equation~\ref{eq:hyperscore}, denoted as $C^*[v]=\argmin_{i\in{1,2,\ldots,p}} w_v^i$. 
As for $c^*[v]$, it is identified by
the assigned cluster of $v$ in the optimal partition $C^*[v]$, according to the weighted $k$-means.
Then we construct the index based on these two variables, denoted as $Index[v]=(C^*[v], c^*[v])$. 

Considering the example shown in Figure~\ref{fig:sp}, we can observe that the optimal partition $C^*[v]$ of the red node $v$ is $C^2$, since the weight of $v$ in $C^2$ (determined by the orange line) is smaller than that in $C^1$. Then in $C^2$, the optimal cluster is $c_1^2$ since the distance between $v$ and $\mu_1^2$ is the minimum among the four centroids. 

\begin{algorithm}[tb!]
\small
\renewcommand{\algorithmicrequire}{\textbf{Input:}}
\renewcommand{\algorithmicensure}{\textbf{Output:}}
\caption{\LocalCEI}
    \begin{algorithmic}[1]
        \REQUIRE Graph $G$, \GNN\ $M$, test set $V_{test}$, query node $v$, set of partitions $P$.  
        \ENSURE Top-$1$ local counterfactual evidence $\LCE_{opt}(v)$ for $v$.  
        
        \STATE $\LCE_{opt}(v) := \texttt{None}$.

        \STATE Identify optimal partition $C^*[v]$ \label{indentify}.
        \STATE Identify optimal cluster $c^*[v]$  \label{select_cls}.

        \FOR{$u \in c^*[v]$}\label{start_local}
            \IF{$M(v)\neq M(u)$ and $\KS(v, u)>\KS(v, \LCE_{opt}(v))$}\label{criteria} 
                \STATE $\LCE_{opt}(v) := u$.
            \ENDIF
        \ENDFOR

        \RETURN $\LCE_{opt}(v)$. \label{end_local}
    \end{algorithmic}
  \label{alg:LocalCEI}
\end{algorithm}
\begin{table}[tb!]
\caption{Statistics of datasets}
\vspace{-2mm}
\label{tab:dataset}
\resizebox{\columnwidth}{!}{
\begin{tabular}{c|cccc}
\hline
dataset & \# nodes & \# edges & \# node features & \# class labels \\ \hline
\german & 1,000 & 22,242 & 27 & 2 \\
\bail & 18,876 & 321,308 & 18 & 2 \\
\cora & 2,708 & 10,556 & 1,433 & 7 \\
\pubmed & 19,717 & 88,648 & 500 & 3 \\
\facebook & 22,470 & 342,004 & 128 & 4 \\ 
\amazon & 1,569,960 & 264,339,468 & 200 & 107 \\ \hline
\end{tabular}
}
\end{table}
\underline{\emph{Querying with the Index.}}
Querying based on the index is straightforward. As detailed in Algorithm~\ref{alg:LocalCEI} (\LocalCEI), given a graph $G$, a \GNN\ $M$, a test set $V_{test}$, a query node $v$, and set of partitions $P$, \LocalCEI\ aims to identify the optimal partition $C^*[v]$ and optimal cluster $c^*[v]$ to query node $v$ (lines~\ref{indentify}-~\ref{select_cls}). 
Inside the optimal cluster $c^*[v]$, \LocalCEI\ applies the same approach as \LocalCEB\ (lines~\ref{start_local}--\ref{end_local}).
To extend the querying process for top-$k$ results, we apply the same approach as \LocalCEB\ within the optimal cluster $c^*[v]$. 

\underline{\emph{Time and Space Complexity.}}
The time cost of \LocalCEI\ consists of four parts, with the first three being similar to those in \LocalCEB: 
{\bf (1)} Inference cost of \GNN: $O\left(Ld\left(|E|+|V_{c}||\mathbf{Y}|\right)\right)$, where $|V_{c}|=\frac{1}{m}|V_{test}|$ indicates the average number of test nodes per cluster; {\bf (2)} vector aggregation cost for \KS\ score: $O(Ld|V_{c}||N(\cdot)|)$, where $|N(\cdot)|$ is the maximum number of 1-hop neighbors of a test node;
{\bf (3)} cost of finding the top-$k$ results: $O(k|V_{c}|+p)$; and {\bf (4)} offline index construction cost: $O(mp|V_{test}|(d+\hs^d))$, where $\hs^d$ is the time for computing the intersection in $d$-dimensional hyperspherical
space~\cite{hyper_inter}.
Therefore, the online time complexity of \LocalCEI\ is roughly reduced by a factor of the number of clusters. 

For space complexity, 
the additional index overhead is $O(p|V_{test}|)$. 

\begin{algorithm}[tb!]
\small
\renewcommand{\algorithmicrequire}{\textbf{Input:}}
\renewcommand{\algorithmicensure}{\textbf{Output:}}
    \caption{\GlobalCE}
    \begin{algorithmic}[1]
        \REQUIRE Graph $G$, \GNN\ $M$, test set $V_{test}$, set of partitions $P$.  
        \ENSURE Top-$1$ global counterfactual evidence $\GCE_{opt}(V_{test})$. 
        \STATE Select \LocalCEB\ or \LocalCEI\ for local \CE\ identification.
        \STATE Identify top-$1$ \LCE\ for each test node based on the selected local algorithm.
        \STATE Identify top-$1$ \GCE\ among the top-$1$ \LCEs.
        \RETURN $\GCE_{opt}(V_{test})$. \label{last}
    \end{algorithmic}
  \label{alg:global}
\end{algorithm}

\subsection{Global \CE\ Identification}
\label{sec:globalCEmethod}
The identification of global \CEs\ is a natural extension of the local algorithms. As shown in Algorithm~\ref{alg:global} , we select one local algorithm (\LocalCEB\ or \LocalCEI) to retrieve the top-$1$ \LCE\ for each test node, and then select the best one as the top-$1$ \GCE. As for the extension to top-$k$ \GCEs, we maintain a bucket $B$ of size $k$, and incrementally add the top-$1$ \LCE\ among all test nodes.

\section{Experimental Results}
\label{sec:exp}

We conduct experiments to demonstrate the effectiveness, efficiency, scalability, and generalizability of our solutions for finding both local and global counterfactual evidences. Our algorithms are implemented in Python 3.10.14 by PyTorch-Geometric framework.  
All experiments are conducted on the single core of a Linux system equipped with AMD EPYC 7302P CPU and 256 GB RAM. 
\textbf{Our code and data are available at~\cite{code}}.
%
\subsection{Experimental Setup}
\textbf{Datasets.}
We utilize datasets from various real-world domains to showcase the performance of our methods. The statistics of the datasets are shown in Table~\ref{tab:dataset}. 

The \german~\cite{nifty} dataset categorizes individuals with good or bad credit risks based on their attributes. It includes features such as gender, loan amount, and account-related details for 1,000 clients.
The \bail~\cite{nifty} dataset contains bail outcome records from various U.S. state courts between 1990 and 2009. It includes past criminal records, demographic details, and other information on 1,8876 defendants released on bail.
The \cora~\cite{planetoid} dataset is a citation network where nodes represent research papers and edges indicate citation links. It contains 2,708 papers categorized into seven topics, with each paper described by a 1,433-dimensional feature vector representing various keywords' presence in the paper text.
\pubmed~\cite{planetoid} is a citation network of medical research papers, where nodes represent articles and edges denote citations. It consists of 19,717 papers from the PubMed database, classified into three categories, with each paper represented by a 500-dimensional feature vector derived from TF-IDF word statistics.
For \facebook~\cite{facebook}, the nodes represent verified pages on Facebook and edges are mutual likes. The node features are extracted from the site descriptions. The task is multi-class classification based on the site category. 
Moreover, we use one large-scale dataset \amazon~\cite{saint} to depict the scalability w.r.t. the number of test nodes. Specifically, for the \amazon\ dataset, the nodes represent the products on the Amazon website and the edges denote the co-purchase by the same customer. The node features are pre-processed features by SVD, and originally were text reviews from the buyer. The task is to classify the product categories. 


\textbf{Graph Neural Networks.} We employ the following \GNNs.  
\emph{Graph convolutional network} (\GCN) is a classic message-passing \GNN~\cite{gcn}.  
\emph{Graph attention network} ({\sf GAT}) dynamically weighs neighbors via attention~\cite{gat}.  
\emph{Graph isomorphism network} ({\sf GIN}) matches the power of the 1-\WL\ test~\cite{gin}.  
\emph{Message passing neural network} ({\sf MPNN}) incorporates learnable edge functions for richer representations~\cite{mpnn}.  
\emph{GraphSAGE} ({\sf SAGE}) extends sampling and aggregating neighbors to handle large graphs efficiently~\cite{graphsage}.

\textbf{Competitors.} To the best of our knowledge, there are no existing methods that can be adapted to the full setting of our problem. Therefore, we compare our index-based solution (Algorithm~\ref{alg:LocalCEI}) with our baseline approach (Algorithm~\ref{alg:LocalCEB}). 
Furthermore, we compare two more variants with missing indexing components, denoted as \LocalCEIwoWC\ and \LocalCEIwoSP. Specifically, \LocalCEIwoWC\ indicates the index algorithm without weighted clustering, while \LocalCEIwoSP\ indicates the index algorithm without supplementary partitioning. 
For the index construction, we set the number of partitions $p=50$ and the number of clusters $m=10$ per partition. The angle $\theta$ for computing \sfield\ is set to $\frac{\pi}{3}$.

\begin{figure}[tb!]
    \centering
    \Description[ftce]{}{}
    
    \subfigure[\pubmed]{\includegraphics[width=0.46 \linewidth]{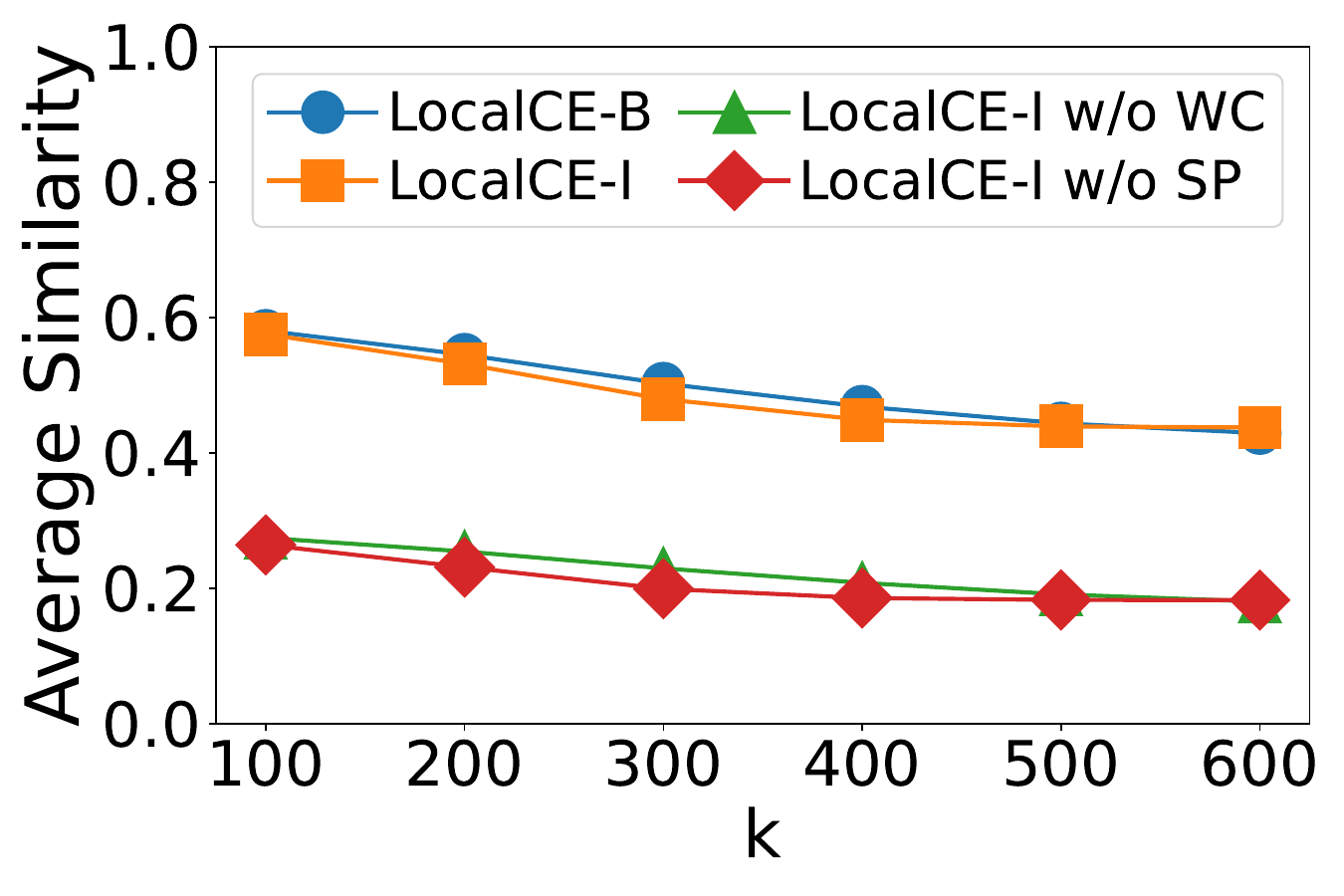}
		\label{fig:pubmed_efcl}}    
    \subfigure[\facebook]{\includegraphics[width=0.46 \linewidth]{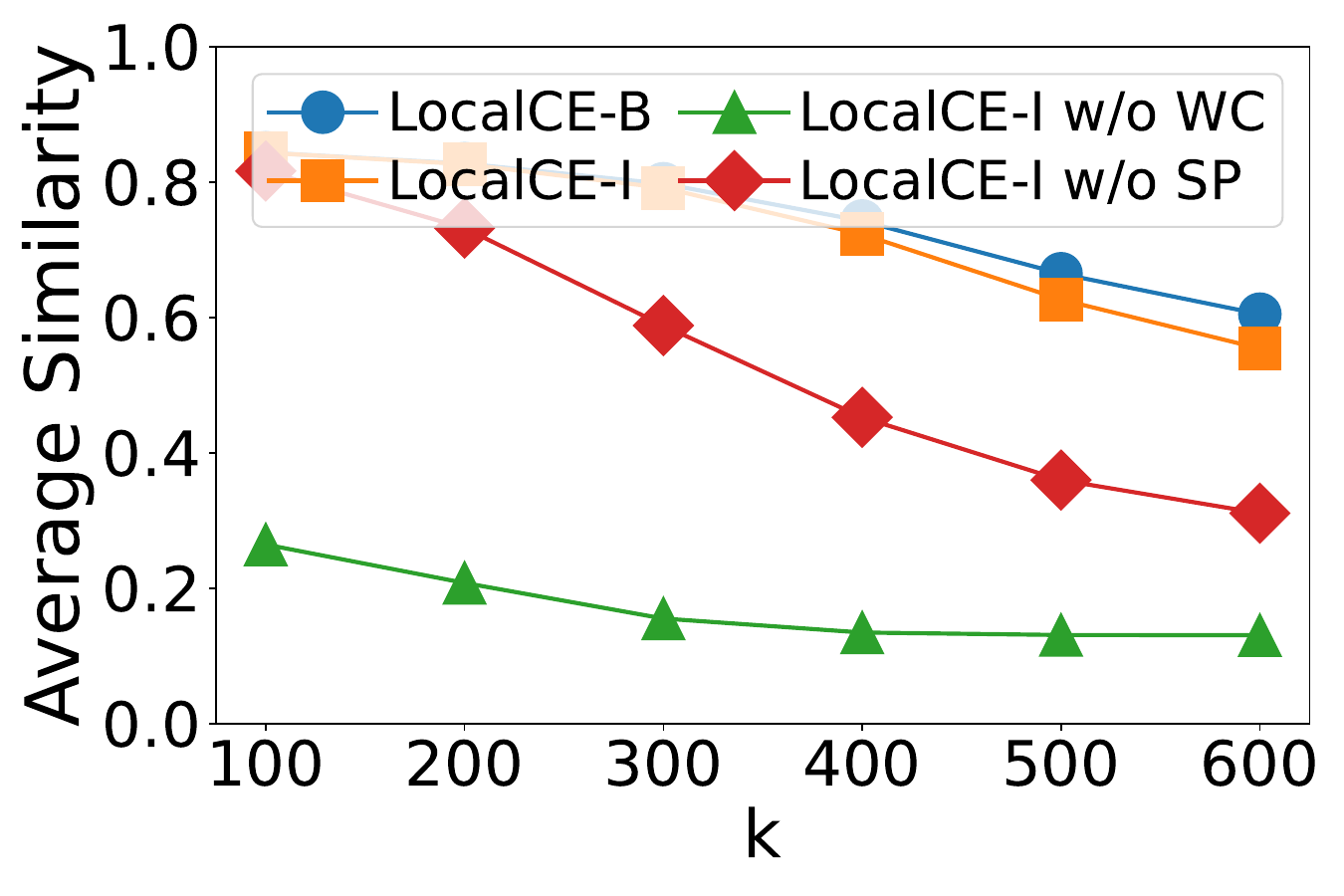}
		\label{fig:facebook_efcl}}   
        
    \subfigure[\pubmed]{\includegraphics[width=0.46 \linewidth]{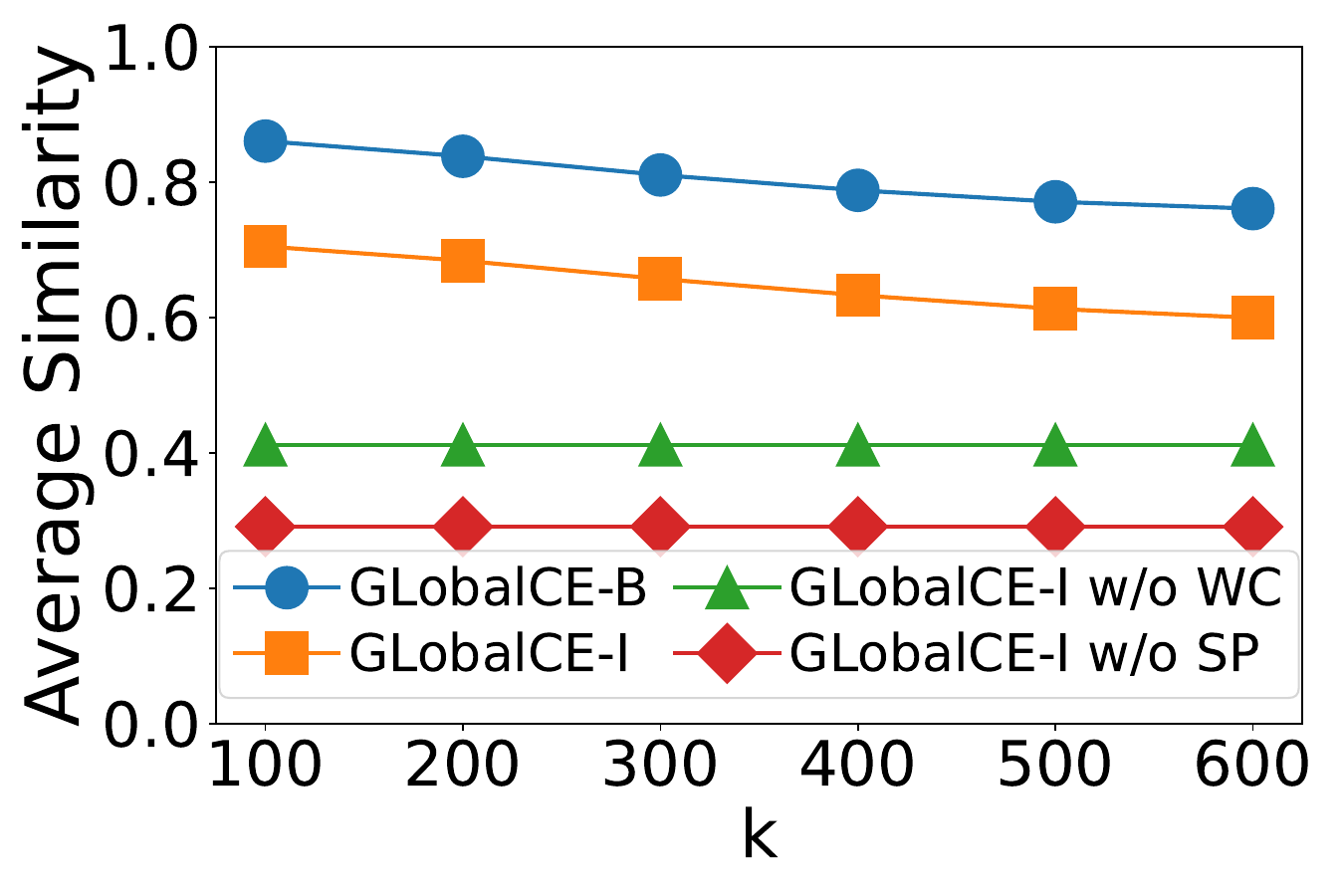}
		\label{fig:pubmed_efcg}} 
    \subfigure[\facebook]{\includegraphics[width=0.46 \linewidth]{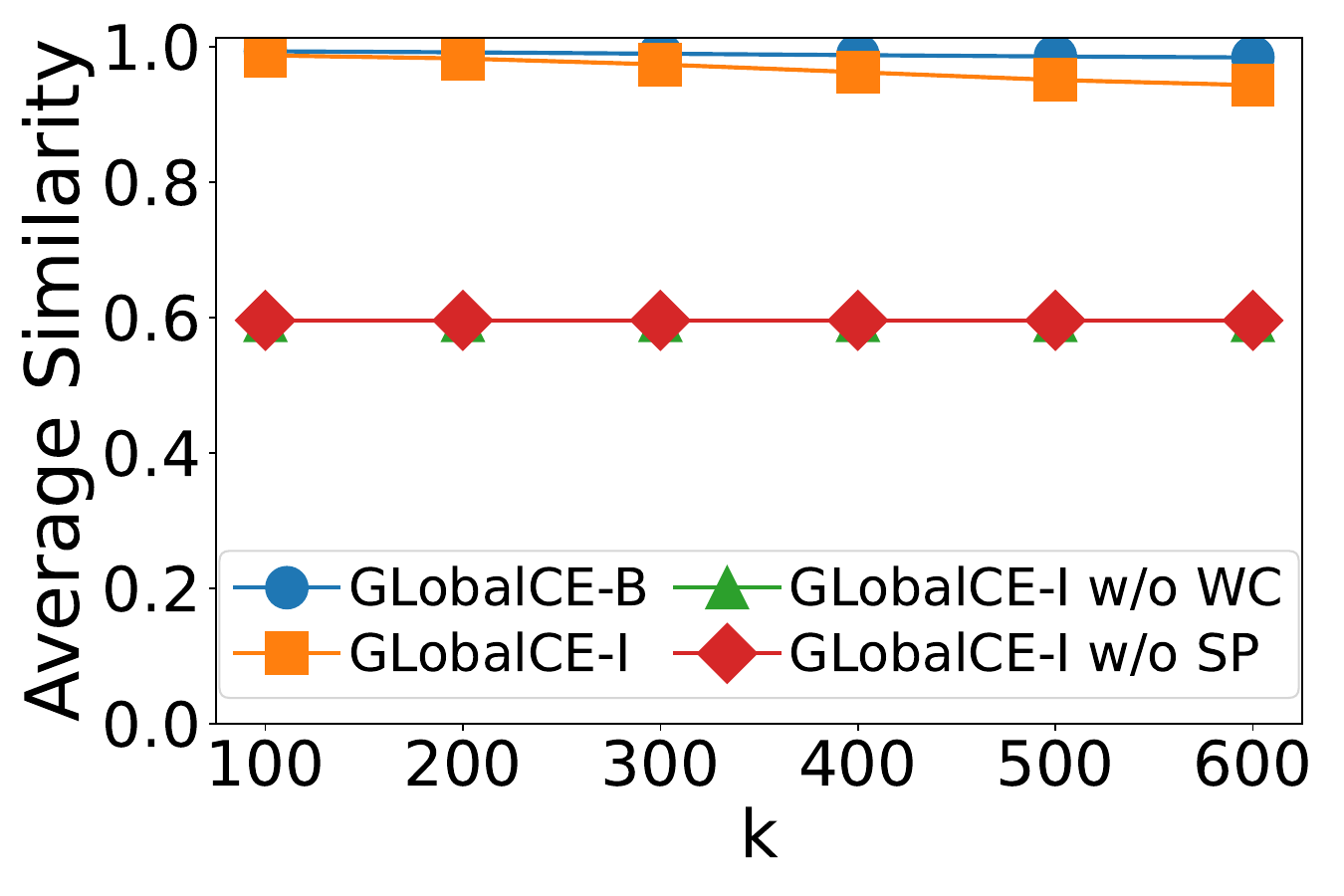}
		\label{fig:facebook_efcg}} 
        
    \subfigure[\pubmed]{\includegraphics[width=0.46 \linewidth]{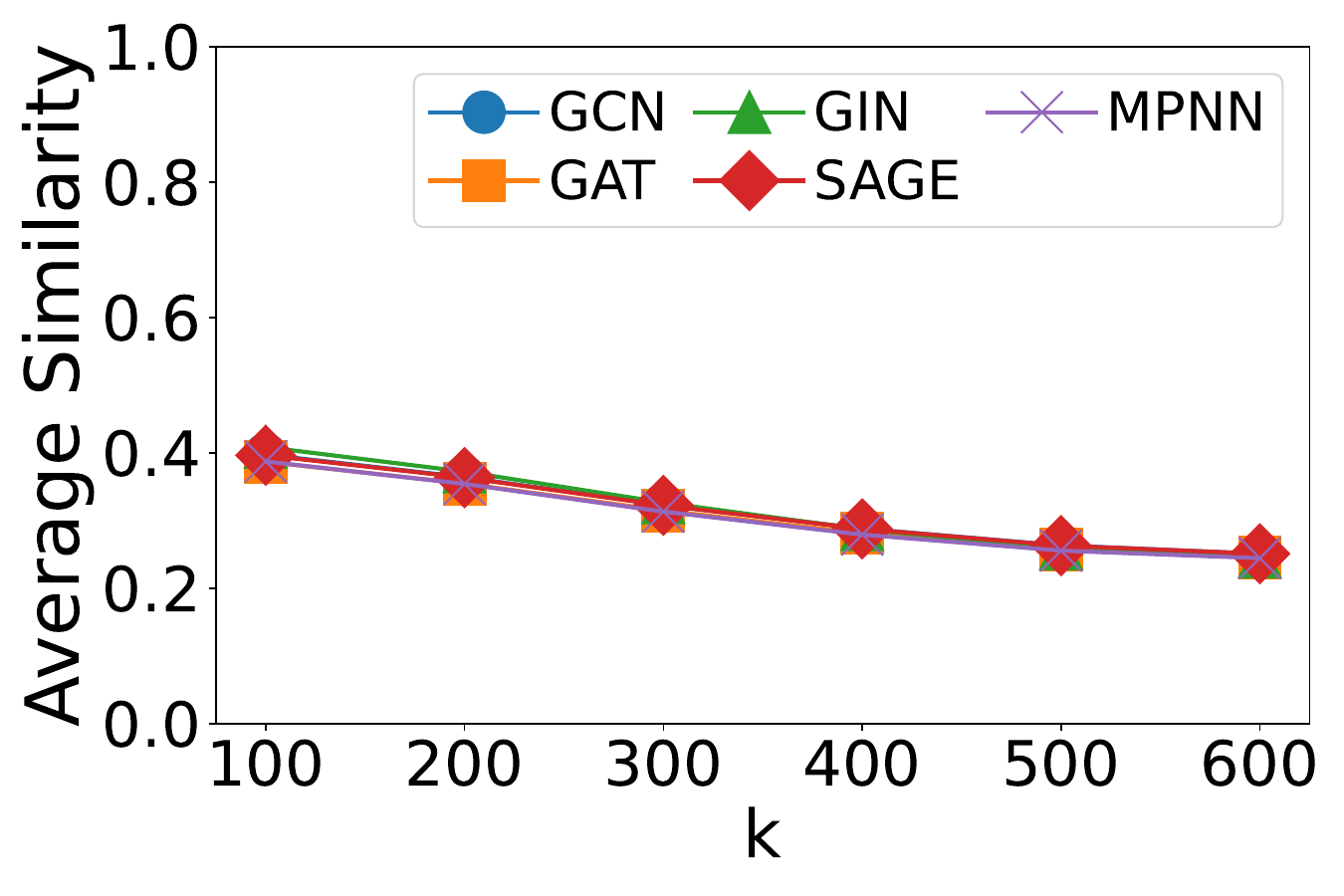}
		\label{fig:pubmed_gnn}}	
    \subfigure[\facebook]{\includegraphics[width=0.46 \linewidth]{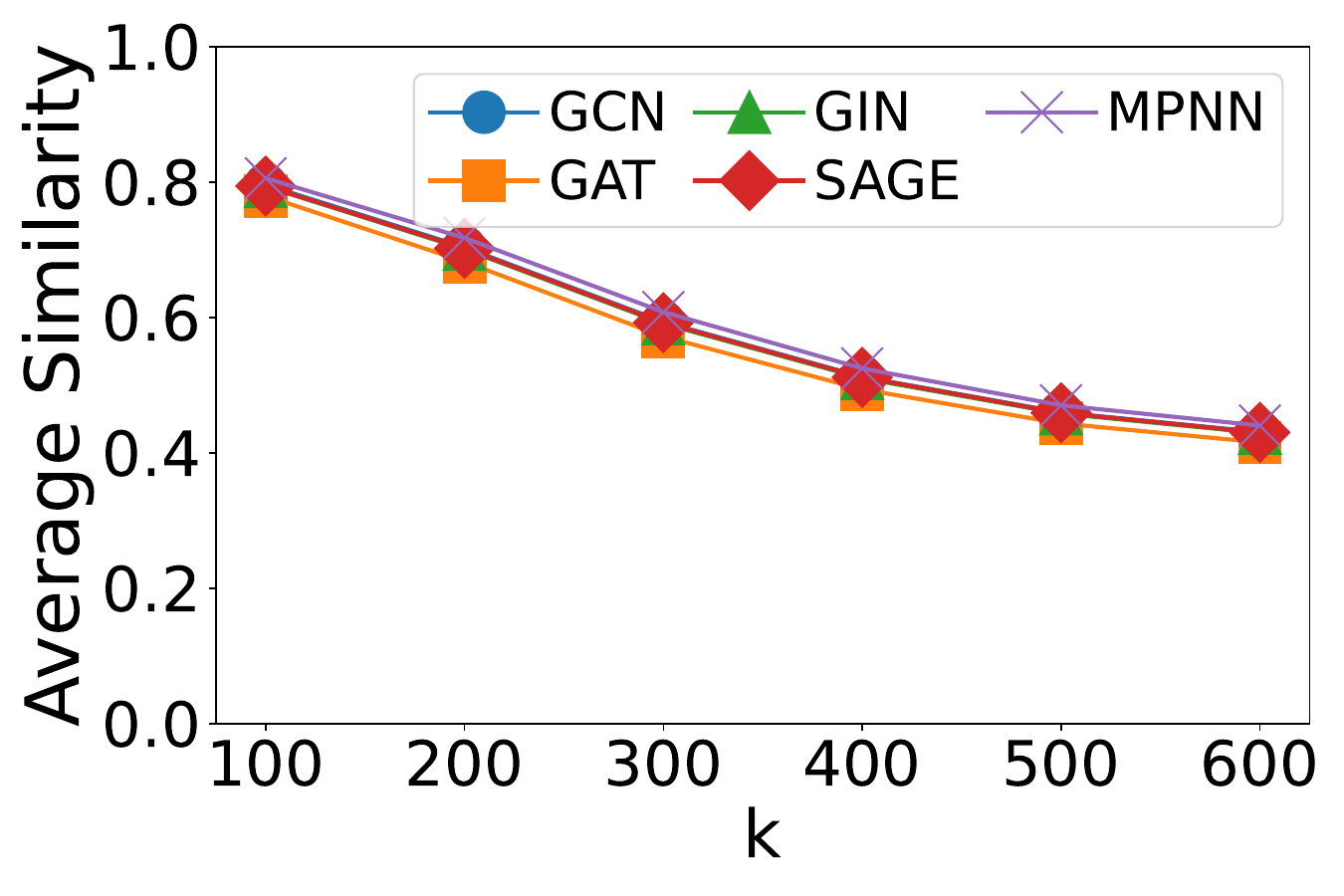}
		\label{fig:facebook_gnn}}	  
    \vspace{-2mm}
    \caption{Effectiveness and generalizability of the proposed algorithms. $k$ denotes the number of top \CEs\ returned.}
    \label{fig:index}
    \vspace{-2mm}
\end{figure}
\begin{figure}[tb!]
    \centering
    \Description[ftce]{}{}

    \subfigure[\facebook]{\includegraphics[width=0.46 \linewidth]{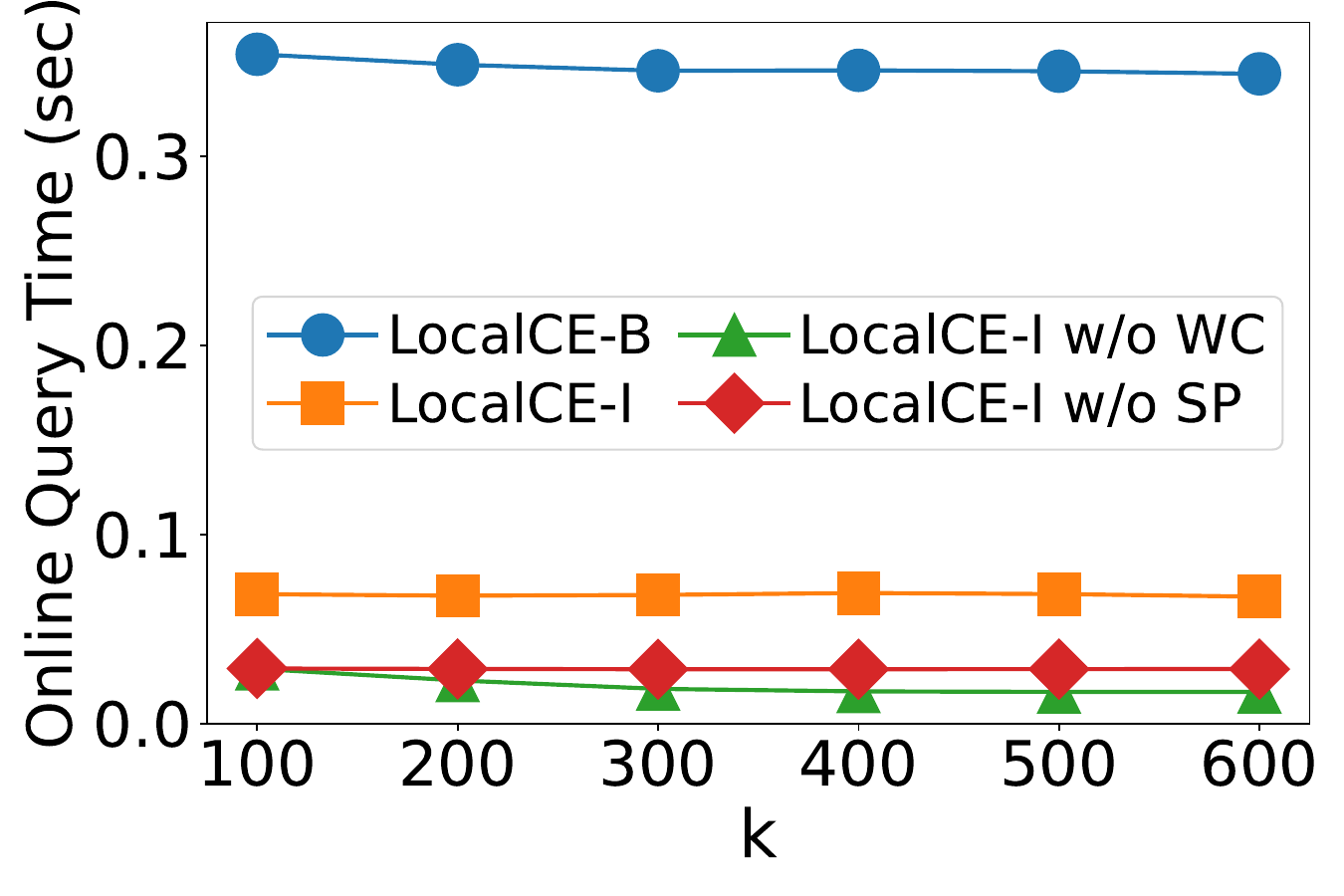}
		\label{fig:facebook_efi}}     
    \subfigure[\amazon]{\includegraphics[width=0.46 \linewidth]{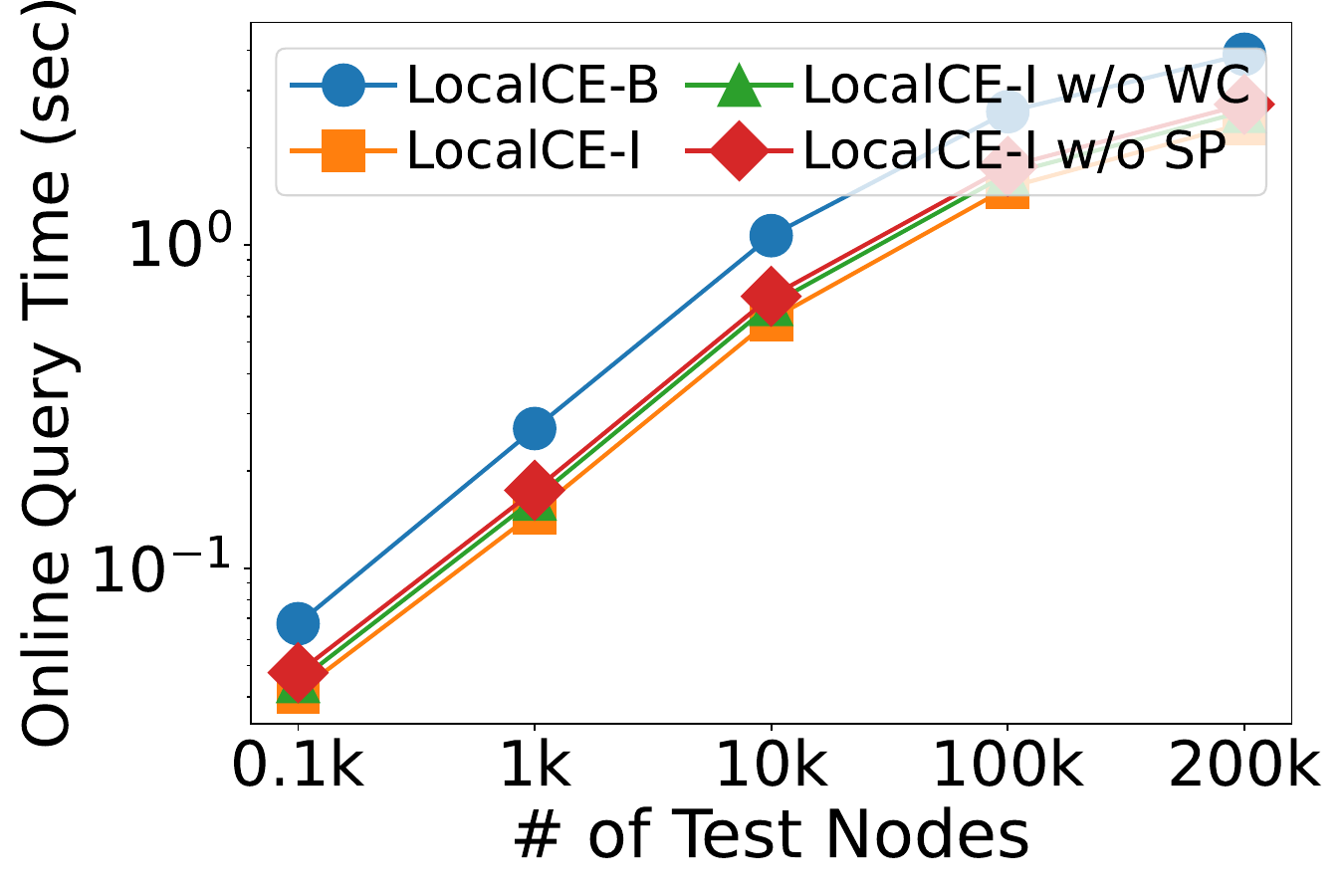}
		\label{fig:amazon_scale}}
    \vspace{-2mm}
    \caption{Efficiency and scalability of the proposed algorithms. $k$ denotes the number of top \CEs\ returned.}
    \label{fig:efi_scal}
    \vspace{-4mm}
\end{figure}

\textbf{Evaluation Metrics.} To evaluate the effectiveness of finding the counterfactual evidences, we use the \emph{average similarity} (\emph{AS}).
\begin{equation}
\small
    AS=\frac{1}{|V_{test}|}\sum_{v\in V_{test}} \frac{1}{k}\sum_{u \in \LCE^k_{opt}(v)} \KS(v,u)
\end{equation}
$\LCE^k_{opt}(v)$ denotes the top-$k$ \LCEs\ of node $v$. Notice that an effective \CE\ finding method would result in higher \emph{AS} score.  

Additionally, we report the running times of our baseline, index creation, and index-based querying algorithms. 
%
\subsection{Effectiveness Results}
Figures~\ref{fig:pubmed_efcl} and~\ref{fig:facebook_efcl} show that for the local top-$k$ counterfactual evidence identification, with the increasing $k$, the average similarity drops. This is because the less promising \LCEs\ have low \KS\ scores than the highly similar \LCEs. Specifically, both the baseline and the index-based algorithm exhibit the capability of identifying high-quality \LCEs, and the effectiveness of \LocalCEB\ and \LocalCEI\ is relatively similar, indicating the advantages of constructing our novel index based on supplementary partitioning and weighted clustering. 
Moreover, the two variants without all indexing components, \LocalCEIwoWC\ and \LocalCEIwoSP, perform worse than our ultimate index-based algorithm \LocalCEI.
This is because neither indexing component alone is sufficient to identify high-quality \CEs\ compared to the baseline algorithm.
Instead, it is the combination of both components that yields results comparable to the baseline search approach.
First, without supplementary partitioning, the weighted clustering approach tends to favor dense vector regions, neglecting boundary nodes and reducing overall similarity among all nodes.
Second, without weighted clustering, supplementary partitioning focuses solely on enhancing local coherence, failing to optimize cluster assignments for the majority of nodes.
Finally, by combining both components, we can assign each node to an appropriate cluster while generating additional clusters for boundary cases. This highlights the effectiveness of our index-based algorithm and underscores the importance of the proposed indexing techniques.

Analogously in Figures~\ref{fig:pubmed_efcg} and~\ref{fig:facebook_efcg}, our global algorithms are capable of identifying the high-quality global top-$k$ counterfactual evidences. Notice that the average similarity in this case is higher than that from local algorithms, since the identified \CEs\ are pair-wise optimal \GCEs. 
Additionally, in the global \CE\ setting, the two variants without indexing components still perform worse than the index-based algorithm. This demonstrates that the indexing techniques effectively identify high-quality \GCEs.
%
\subsection{Generalization to Different \GNNs}
We retrieve global counterfactual evidences (\GCEs) considering various state-of-the-art \GNNs\ to explore the predictive behavior among different \GNNs\ and to demonstrate how well our \GCE\ finding algorithms generalize across different \GNNs. As given in Figures~\ref{fig:pubmed_gnn} and~\ref{fig:facebook_gnn}, different models show similar trends on identifying \CEs, with slight difference in the average similarity. Such results indicate that the counterfactual evidences returned by our methods are model-agnostic to different message-passing \GNNs. 
\begin{table}[tb!]
    \centering
    \small
    \vspace{-2mm}
    \caption{Node features and their discrimination scores (DS) considering \GCN\ \cite{gcn} and \fairgnn\ \cite{fairgnn}: {\em Bail} dataset.}
    \label{tab:discrimination_scores}
    \vspace{-2mm}
    \begin{tabular}{p{1.8cm}|p{1.2cm}}
        \hline
        \multicolumn{2}{c}{\GCN} \\
        \hline
        \textbf{Features} & \textbf{$DS$} \\
        \hline
        WHITE & 0.55 \\
        WORKREL & 0.52 \\
        FILE3 & 0.52 \\
        SUPER & 0.46 \\
        FILE1 & 0.41 \\
        MARRIED & 0.37 \\
        FILE2 & 0.35 \\
        FELON & 0.33 \\
        SCHOOL 8-13 & 0.32 \\
        PROPTY & 0.27 \\
        \hline
    \end{tabular}
    \quad
    \begin{tabular}{p{1.8cm}|p{1.2cm}}
        \hline
        \multicolumn{2}{c}{\fairgnn} \\
        \hline
        \textbf{Features} & \textbf{$DS$} \\
        \hline
        TIME 0-30 & 0.79 \\
        TIME > 30 & 0.79 \\
        SUPER & 0.52 \\
        FELON & 0.51 \\
        FILE3 & 0.48 \\
        WHITE & 0.48 \\
        WORKREL & 0.48 \\
        MARRIED & 0.46 \\
        FILE2 & 0.44 \\
        PROPTY & 0.42 \\
        \hline
    \end{tabular}
\end{table}
%
\subsection{Efficiency and Scalability Results}
Figure
~\ref{fig:facebook_efi} presents the online query times of \LocalCEB, \LocalCEI, \LocalCEIwoWC, and \LocalCEIwoSP -- they are generally less sensitive to the parameter $k$ (i.e., the number of top \CEs\ returned). Such results show the advantage of the proposed algorithms for fast identification of counterfactual evidences, which will provide swift results for downstream applications. Meanwhile, for all three index-based algorithms, the online query time is significantly faster than \LocalCEB. The proposed index construction approach efficiently prunes dissimilar nodes, and meanwhile maintains sufficient performance. The index construction overhead is $159$ sec and $358$ sec on \pubmed\ and \facebook\ datasets, respectively. 
As for the scalability, we show online query time on the million-scale dataset \amazon. We observe from Figure~\ref{fig:amazon_scale} that the online query time of \LocalCEB\ linearly increases with the number of test nodes. 
Meanwhile, the online query time of all three index-based algorithms also increases with the number of test nodes, but is always lower than that of \LocalCEB, since the desired counterfactual evidences are identified from the cluster determined by the index. 
The index construction requires $1503$ sec on \amazon. 

\section{Applications} 
\label{sec:applications}
Various downstream tasks can benefit from utilizing our counterfactual evidences. We demonstrate the effectiveness and insights they bring to these tasks by showcasing the following applications.
%
\subsection{Revealing Unfairness of \GNNs}
\label{app:fair}
\begin{figure}[tb!]
    \centering
    \Description[vpe]{}{}
    \subfigure[Cora]{\includegraphics[width=0.47 \linewidth]{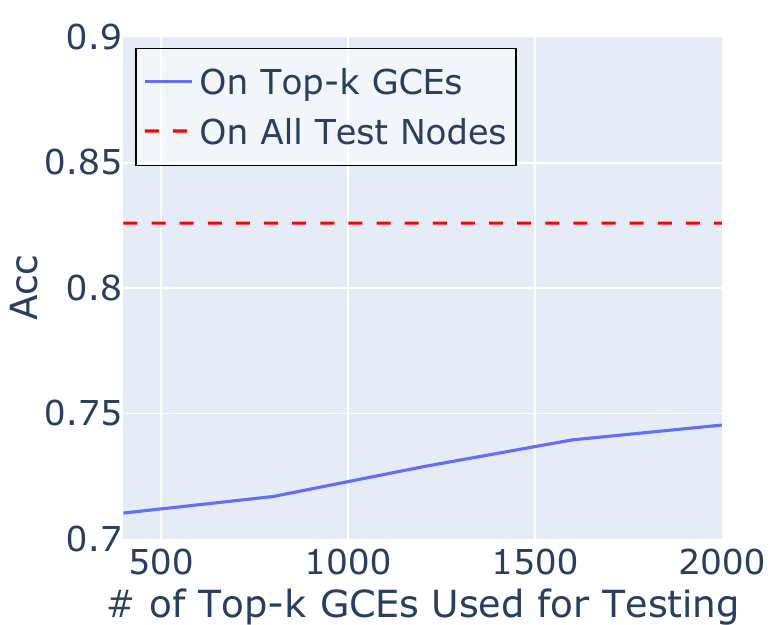}
		\label{fig:vpe_Cora}}    
    \subfigure[PubMed]{\includegraphics[width=0.47 \linewidth]{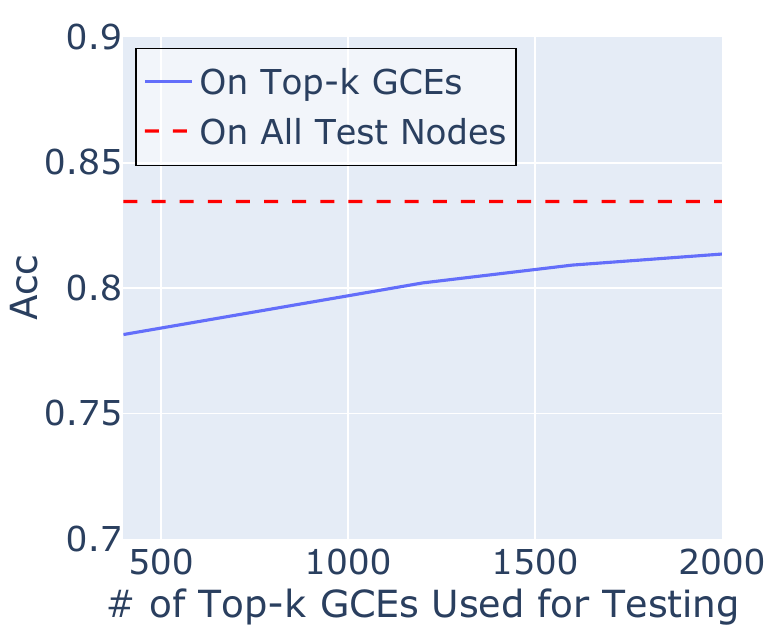}
		\label{fig:vpe_PubMed}}	
    \vspace{-2mm}
    \caption{Accuracy within the top-$k$ Global \CEs: Accuracy within the \GCEs\ is significantly lower for smaller $k$,
    mainly consisting of borderline nodes, which are difficult for the \GCN\ to classify correctly.}
    \label{fig:vpe}
    \vspace{-2mm}
\end{figure}
\begin{figure}[tb!]
    \centering
    \Description[ftce]{}{}

    \subfigure[PubMed]{\includegraphics[width=0.47 \linewidth]{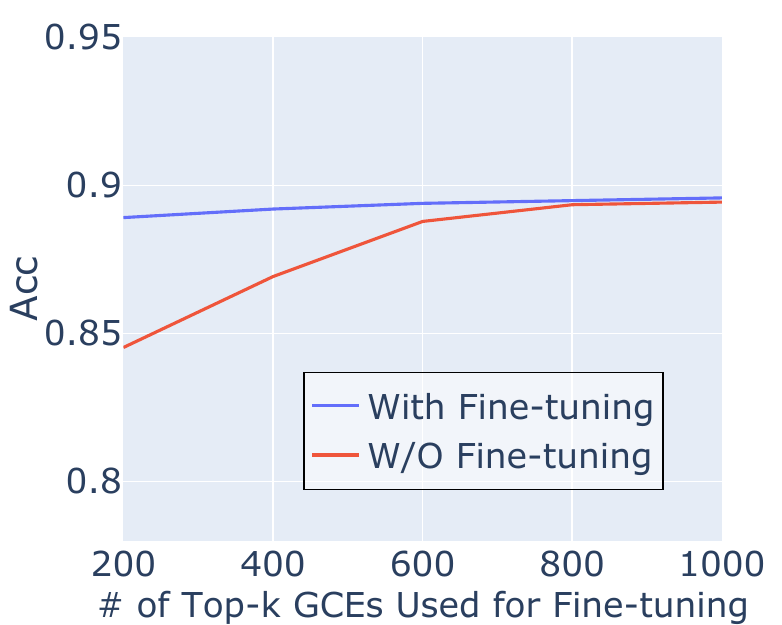}
		\label{fig:2_ftce_PubMed}}	
    \subfigure[PubMed]{\includegraphics[width=0.47 \linewidth]{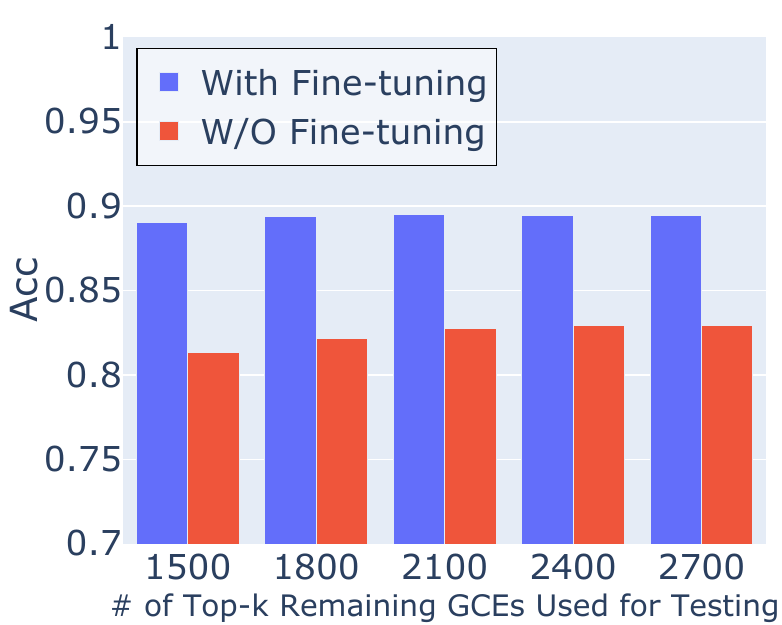}
		\label{fig:ftce_PubMed}}
    \vspace{-2mm}
    \caption{(a): Fine-tuning the \GCN\ model with a set of top-$k$ \GCEs\ as a validation set improves the accuracy on the remaining test nodes. (b): We fix the top-1200 \GCEs\ as a validation set to fine-tune the \GCN, which not only improves the accuracy on the remaining test nodes, but the accuracy also remains consistently high after fine-tuning on both borderline and relatively easier cases. 
    }
    \label{fig:ftce}
     \vspace{-2mm}
\end{figure}
Ensuring fairness in \GNNs\ promotes ethical decision-making by preventing biases related to sensitive node features such as gender and race, particularly in  scenarios like credit defaulter identification~\cite{credit} 
and court trial decisions~\cite{bail}. 
We will use \CEs\ to detect whether a \GNN\ model is fair, which is crucial for ensuring that the model's decisions do not perpetuate biases and inequalities.

To measure node feature importance, we introduce the notion of {\em discrimination score} for a node feature value $f()=f_i$ at a test node $v$, denoted as $DS\left(f(v)=f_i\right)$. In particular, we consider the top-$k$ local \CEs\ of the test node $v$, denoted by $\LCE_k(v)$, and compute.
\begin{equation}
\small
    DS\left(f(v)=f_i\right) = \frac{1}{k}\sum_{u \in \LCE_k(v)} \mathbb{I}\left(f(v)=f_i \wedge f(u)\ne f_i\right)
    \label{eq:DiscrimScore}
\end{equation}
$\mathbb{I}()$ is an indicator function which ensures that if a feature value $f()=f_i$ occurs at node $v$ and the same feature value does not occur frequently in $v$'s top-$k$ local \CEs, then $f(v)=f_i$ is crucial towards $v$'s predicted class label. We set $k=10$ in our experiments.

We employ a state-of-the-art fair GNN model, \fairgnn~\cite{fairgnn}, as our classifier. Meanwhile, we employ \GCN\ as the classic GNN model for comparison. We apply these models to the \bail\ dataset.
We observe from Table~\ref{tab:discrimination_scores} that the \GCN's prediction for bail decisions heavily depends on the sensitive feature ``race=WHITE'',
with a particularly high $DS$ when using our method based on \CEs. 
In contrast, \fairgnn\ reduces this dependency on the sensitive feature, with ``race=WHITE'' dropping to the sixth position in importance, which means that \fairgnn\ can achieve fairer predictions by mitigating racial bias in its decision-making process.
These results demonstrate the superiority of \CEs\ in detecting fairness in \GNNs.
%
\subsection{Verifying Prediction Errors}
\label{app:error}
\GNNs\ are prone to producing prediction errors, particularly when dealing with borderline cases---nodes that lie near the decision boundary of \GNNs~\cite{borderline}. 
Verifying these cases can reveal underlying issues and guide improvements for \GNNs, offering opportunities for enhancing model robustness and developing more effective downstream applications. We will explore how to use \CEs\ to verify prediction errors in \GNNs.

We employ the classic \GCN\ as the classifier and use two widely used graph node classification datasets: \cora~\cite{cora} and \pubmed~\cite{pubmed}. 
We evaluate the effectiveness of \CEs\ based on the accuracy of the classifier, where the accuracy is the ratio of the number of correctly classified nodes to the total number of nodes.

In Figure~\ref{fig:vpe}, we show the accuracy across all test nodes (red dashed lines) and within the top-$k$ Global \CEs\ (\GCEs, blue lines). The accuracy within the \GCEs\ is significantly lower for smaller values of $k$, which indicates that at lower $k$ values, the \GCEs\ are dominated by borderline nodes that are difficult for the \GCN\ to classify correctly. As the value of $k$ increases, the borderline nodes become a smaller fraction of the overall \GCEs\ considered, reducing their negative impact on classification accuracy, 
as reflected by the blue line approaching the red dashed line. 
%
\subsection{Fine-tuning with Counterfactual Evidences}
\label{app:finetune}
%
In this section, we utilize a limited number of \CEs\ to 
improve the  prediction accuracy of \GNNs. 
Similar to the previous application, we choose the classic \GCN\ as the classifier and use the 
\pubmed\ datasets. 
We use a set of top-$k$ \GCEs\ as a validation set to fine-tune the \GCN\ model and compare the classification accuracy of the model before and after fine-tuning, on the remaining test nodes.

Figure~\ref{fig:2_ftce_PubMed} shows the overall performance before and after fine-tuning. We observe that the performance improvement is significant at the beginning since borderline nodes are predominant in the \GCEs. 
With the increasing number of \GCEs, the performance gap narrows, since the models are not significantly different in the remaining easier test nodes (i.e., non-borderline nodes). 

Next, we fix the top-1200 \GCEs\ as a validation set to fine-tune the \GCN\, 
and depict the classification accuracy on varying numbers of remaining top-$k$ \GCEs. Figure \ref{fig:ftce_PubMed} demonstrates that the model achieves substantial performance improvement after fine-tuning using the top-1200 \GCEs\ as a validation set. This is because the borderline nodes identified by \GCEs\ are incorporated into the model. Additionally, after fine-tuning, the
classification accuracy remains consistently high on both borderline and relatively easier cases, which is evident as we test with varying numbers of remaining top-$k$ \GCEs. Meanwhile,  before fine-tuning, the model was more accurate for relatively easier instances.

\section{Conclusions}
\label{sec:conclusions}
We proposed \emph{Counterfactual Evidences} for node classification, a novel concept that identifies node pairs with high structural and feature-based similarity, yet predicted differently by a \GNN. We studied both local and global variants of the problem, by transforming them to the $k$-nearest neighbor search using cosine similarity over high-dimensional, dense vector spaces, and developed 
high-quality index-based algorithms with two new concepts: supplementary partitioning and weighted clustering. We conducted experiments to illustrate the effectiveness, efficiency, generalizability, and scalability of our solution. 
Then, we showcased counterfactual evidences' capability in revealing model unfairness, verifying prediction errors, and the capacity to effectively fine-tune \GNNs.
In future, we shall study counterfactual evidences for other graph machine learning problems. 

\begin{acks}
Dazhuo Qiu and Arijit Khan acknowledge support from the Novo Nordisk Foundation grant NNF22OC0072415. 
Arijit Khan and Yan Zhao are corresponding authors.
\end{acks}

\bibliographystyle{ACM-Reference-Format}
\bibliography{ref}
\appendix


\section{Additional Case Studies}
\label{sec:additional_case_study}

We recall that the existing ``\textit{counterfactual} \textbf{explanation generation}'' problem 
\cite{LucicHTRS22, BajajCXPWLZ21, numeroso2021, MaGMZL22} and our proposed ``\textit{counterfactual} \textbf{evidence search}'' problem are fundamentally different (\S \ref{sec:introduction}).  In this section, we additionally present several case studies to highlight that the state-of-the-art counterfactual explanations could often suggest ``infeasible'' perturbations in reality, thus it might be difficult to take recourse actions based on the generated counterfactual explanations. In contrast, our proposed counterfactual evidence identifies a ``real'' instance already present in the dataset, which is quite similar to the query instance, yet classified differently by the pre-trained \GNN. 
Therefore, we can investigate the issues with the \GNN, such as unfairness (\S~\ref{app:fair}) and prediction errors (\S~\ref{app:error}), as well as debug them, e.g., fine-tuning the \GNN\ 
(\S~\ref{app:finetune}), in a more practical manner based on the real evidences. 
The following case studies showcase the differences between counterfactual explanation and counterfactual evidence, while spanning diverse domains such as biology, finance, and law. 

\subsection{Case Study - 1}
\begin{figure}[t]
    \centering
    \Description[protein example]{}{}
    \includegraphics[width=0.96\linewidth]{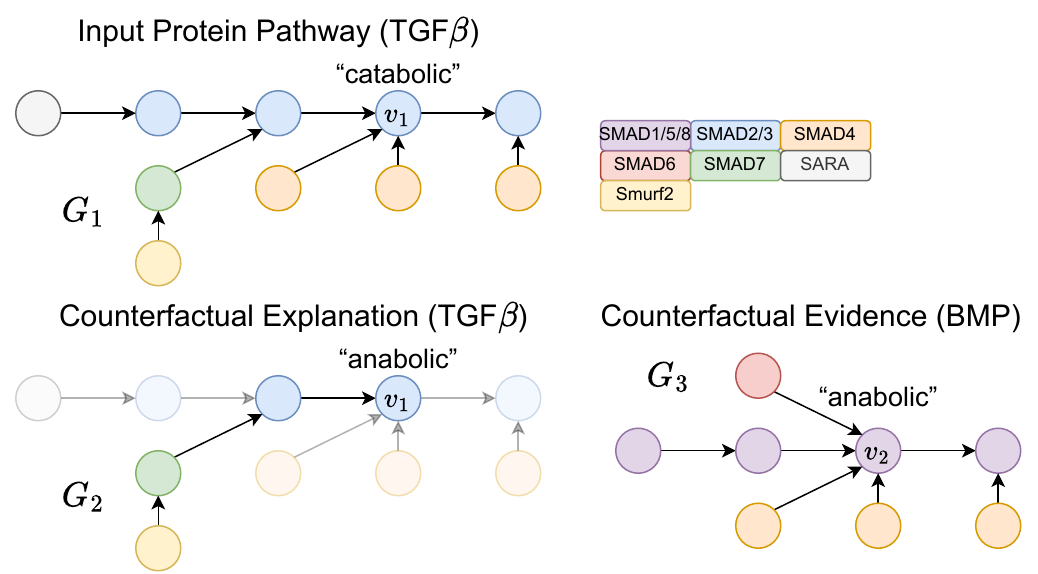}
    \caption{\small Each node is a protein and the directed edges indicate the interactions. 
    Node $v_1$ is associated with a graph $G_1$, denoted as a protein pathway $TGF\beta$. $v_1$ is classified as “catabolic”.
    $G_2$ is a counterfactual explanation, which is a fraction of the $TGF\beta$ pathway graph $G_1$.
    $G_3$ is the associated graph of counterfactual evidence $v_2$, denoted as another protein pathway $BMP$. $v_2$ is classified as “anabolic”.
    }
    \label{fig:protein_example}
\end{figure}
Consider the following example, as shown in Figure~\ref{fig:protein_example}.
Each node is a protein, and the directed edges indicate the interactions between proteins. The node classification task is to predict whether the node belongs to “catabolic” or “anabolic”. 
Node $v_1$ is associated with a graph $G_1$, denoted as a protein pathway $TGF\beta$.
The $TGF\beta$ pathway promotes catabolic processes, leading to cartilage degradation~\cite{proteinpathway}; thus, $v_1$ is classified as “catabolic.”

$G_2$ is a counterfactual explanation, which is a fraction of the $TGF\beta$ pathway graph $G_1$. 
If we keep it but remove the rest of $G_1$ (faded nodes and edges), then $v_1$ will be classified differently as “anabolic”. 
However, a fraction of the $TGF\beta$ pathway is not functional in reality, as the pathway's functionality relies on the inclusion of all essential interactions. Each path within the pathway must be present to ensure its proper activation and biological effect. Therefore, such an invalid structure could be less practical to domain experts and for the analysis of articular cartilage homeostasis.

$G_3$ is the associated graph of counterfactual evidence $v_2$, denoted as another protein pathway $BMP$.
The $BMP$ pathway promotes anabolic processes by stimulating cartilage matrix production~\cite{proteinpathway}; thus, $v_2$ is classified as “anabolic.”
$G_3$ is a functional protein pathway and shares a similar structure with $G_1$. Meanwhile, $TGF\beta$ and $BMP$ play opposite roles in articular cartilage homeostasis~\cite{proteinpathway}. 
From this, we can intuitively observe that protein types are crucial for the functionality of pathways:
When the pathway primarily consists of ‘SMAD2/3’ proteins, it is likely to promote the catabolic process, leading to cartilage degradation.
In contrast, if the pathway primarily consists of ‘SMAD1/5/8’ proteins, it is likely to promote the anabolic process: cartilage growth and repair.

\subsection{Case Study - 2} 


We tested GNNExplainer on both Bail and German datasets and found that two sensitive attributes “race” and “gender”, respectively, are identified as critical features whose values should be perturbed to get a different prediction on test nodes. Indeed, we verified that 2947 test instances on Bail and 61 test instances on German would show different predictions after perturbing their sensitive features. However, (1) in reality, it is difficult or even impossible to change such sensitive features for a real human user. (2) Besides, the “newly” generated instances after such perturbation do not actually exist in the original datasets. 

For example, customer 223 in the German dataset is a “male” customer predicted as a “good customer” (i.e., will get a loan). When changing this customer’s gender to “female”: (1) the “new” test instance will be predicted as a “bad customer”; (2) however, such an update of gender is difficult to often impossible for a real human customer; and (3) such a “newly” generated hypothetical “female” customer does not actually exist in the German dataset. Similarly, defendant 165 in the Bail dataset is a “white” person and is predicted as “receiving bail”, when changing the defendant’s race to “non-white”: (1) the user is predicted as “not receiving bail”; (2) however, such update of race is generally impractical in reality, and (3) such a “newly” generated hypothetical “non-white” defendant does not exist in the Bail dataset. 

This shows that infeasible cases are common issues for existing GNN Explanation methods and our methods address this problem effectively by identifying counterfactual nodes without “hypothetical” perturbations.

\end{document}